\documentclass[nohyperref]{article}

\usepackage{microtype}
\usepackage{graphicx}
\usepackage{booktabs} 
\usepackage{subcaption}

\usepackage{hyperref}

\usepackage[accepted]{icml2022}

\usepackage{amsmath}
\usepackage{amssymb}
\usepackage{mathtools}
\usepackage{amsthm}

\usepackage[capitalize,noabbrev]{cleveref}
\usepackage{balance}
\usepackage{setspace}
\usepackage{stfloats}
\usepackage{framed}

\theoremstyle{plain}

\theoremstyle{definition}

\theoremstyle{remark}

\usepackage[textsize=tiny]{todonotes}

\icmltitlerunning{Physics-constrained Unsupervised Learning of Partial Differential Equations using Meshes}

\begin{document}

\twocolumn[
\icmltitle{Physics-constrained Unsupervised Learning of Partial Differential Equations using Meshes}


\icmlsetsymbol{equal}{*}

\begin{icmlauthorlist}
\icmlauthor{Mike Y. Michelis}{yyy}
\icmlauthor{Robert K. Katzschmann}{yyy}
\end{icmlauthorlist}

\icmlaffiliation{yyy}{Soft Robotics Lab, ETH Zurich, Switzerland}

\icmlcorrespondingauthor{Mike Y. Michelis}{michelism@ethz.ch}

\icmlkeywords{Machine Learning, ICML, physics-based, physics-informed, unsupervised, physics, PDE, mesh, FEM, irregular grid}

\vskip 0.3in
]



\printAffiliationsAndNotice{\icmlEqualContribution} 

\begin{abstract}

Enhancing neural networks with knowledge of physical equations has become an efficient way of solving various physics problems, from fluid flow to electromagnetism. Graph neural networks show promise in accurately representing irregularly meshed objects and learning their dynamics, but have so far required supervision through large datasets.
In this work, we represent meshes naturally as graphs, process these using Graph Networks, and formulate our physics-based loss to provide an unsupervised learning framework for partial differential equations (PDE). 
We quantitatively compare our results to a classical numerical PDE solver, and show that our computationally efficient approach can be used as an interactive PDE solver that is adjusting boundary conditions in real-time and remains sufficiently close to the baseline solution. Our inherently differentiable framework will enable the application of PDE solvers in interactive settings, such as model-based control of soft-body deformations, or in gradient-based optimization methods that require a fully differentiable pipeline.



\end{abstract}

\section{Introduction}
\label{sec:intro}


\begin{figure}[htb]
\centering
\includegraphics[width=\columnwidth]{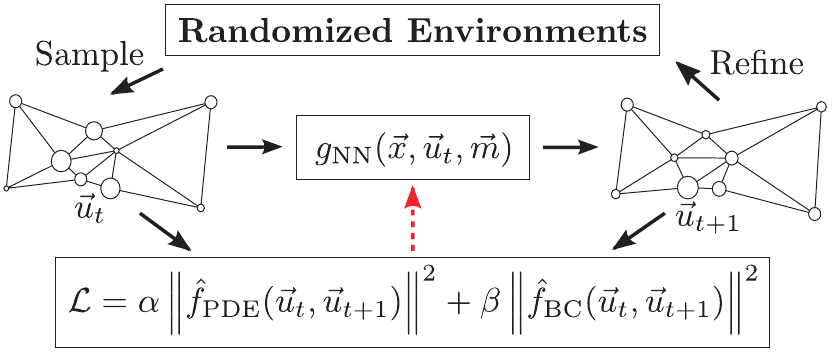}
\caption{The graphical representation shows the interaction of the physics-constrained loss, the graph network input/output, and the environment randomization during training. The dataset is initialized with random initial conditions, such as randomly placed obstacles or varied inflow conditions. Random minibatches are sampled during training, and the predictions $\vec{u}_{t+1}$ are added back into the dataset for the next timestep. Using $\vec{u}_{t}$ and $\vec{u}_{t+1}$, the graph network $g_{\mathrm{NN}}$ is trained based on the shown physics-constrained loss. Besides the position $\vec{x}$ and solution $\vec{u}_t$, the network $g_{\mathrm{NN}}$ also receives $\vec{m}$ as input, which represent extra nodal information such as node types, material parameters, \emph{etc.}}
\label{fig:outline}
\end{figure}

\begin{figure}[htb]
\centering
\includegraphics[width=\columnwidth]{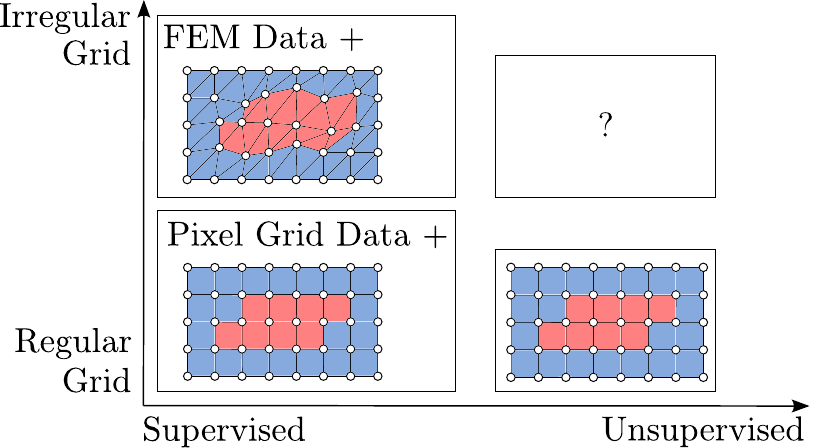}
\caption{Existing physics-constrained approaches have either used unsupervised data generation with regular grids, or supervised data generation on regular or irregular grids. Unsupervised data generation means that instead of needing ground-truth data that is expensive to generate, one uses randomly generated environments to train the network based on a physics-constrained loss. While regular grids apply the same resolution everywhere, irregular grids can accurately represent varying levels of details for a given problem. We propose in this work to tackle the missing piece that combines the best of both worlds: unsupervised data generation with irregular grids.}
\label{fig:perspective}
\end{figure}

Partial Differential Equations (PDE) govern all of physics, yet in many practical scenarios with complex boundary conditions, such as the Navier-Stokes equations for fluid flow or Maxwell's equations for electromagnetism, finding analytical solutions is infeasible. In practice, these PDEs are solved using numerical methods such as finite difference, finite volume, or finite element method. Though accurate, these methods can be computationally expensive. Especially in settings where a large number of PDE simulations are required, for instance shape optimization of solids immersed in fluids, the use of direct numerical methods becomes prohibitive \citep{ryzhakov2017fast}.

As deep learning has grown in popularity, we have seen plenty of works applying deep neural networks solving hard PDEs, the most popular being in the domain of fluid mechanics. Due to the success of Convolutional Neural Networks (CNN), the problem is solved by representing the PDE solution as images on regular grids where every pixel covers a same-sized area \citep{guo2016convolutional, tompson2017accelerating, thuerey2020deep, wandel2020learning}. The limitation with these regular grids, however, is their poor ability to handle complex, irregular geometries. They fail to adaptively refine and coarsen the resolution based on how accurate solutions are in certain positions on the grid. Additionally, only Eulerian simulations (fixed grid) are permissible under the CNN formulation, hence an alternative is necessary for problems that require a Lagrangian simulation method where the grid itself moves, \emph{e.g.}, hyperelastic solid deformations \citep{luo2018nnwarp}.

Instead of using regular grids, there has been a growing interest in solving PDEs using point-cloud-based inputs that can achieve more accurate solutions near complex geometries by increasing sampling in those areas \citep{raissi2019physics, sanchez2020learning, kashefi2021point}. These point clouds are naturally represented using graphs; \citet{sanchez2020learning} have applied their Graph Network (GN) framework \citep{sanchez2018graph} to process these point-cloud inputs to better include vertex-to-vertex interactions. Polygonal meshes are another representation that can appropriately account for irregular geometries; \citet{pfaff2021learning} have used the same GN framework to successfully learn a variety of physical phenomena on triangular mesh inputs, speeding up the simulation by up to 290x, while also showing that they outperform particle- and grid-based baselines.

Most of the previous approaches employ supervised learning, requiring a large training dataset from expensive ground-truth numerical solvers. Instead, a new trend \citep{karniadakis2021physics} shows how we can leave behind this big-data driven learning, and solve PDEs in an unsupervised manner by constraining solutions of the network to those that uphold the governing physical equations. We hypothesize that supervised methods that have no knowledge of the underlying physical equations do not generalize well, since they have not understood the underlying phenomena. More often than not, PDEs require higher-order derivatives of solution variables, which Physics-Informed Neural Networks (PINN) \citep{raissi2019physics, raissi2018hidden, haghighat2020deep, mishra2021physics} solve by leveraging the autodifferentiability of neural networks. Unfortunately, PINNs are trained with one fixed boundary condition, and have no concept of vertex neighborhoods, hence adding boundary conditions like static obstacles at test-time will fail. Unlike PINNs, \citet{wandel2020learning, wandel2021spline} build the physics-constrained loss specifically for incompressible fluid dynamics using finite difference gradients, as they work on regular grids using CNNs to process the network solution. By using such images in combination with CNNs, and including boundary conditions in the network input, \citet{wandel2020learning} manage to handle interactive obstacles at test-time.

While PINNs struggle with generalizing to variable boundary conditions, \citet{wandel2020learning} is unable to incorporate accurate geometries due to the uniform grid resolution of images. In this work, we will combine the representational power of meshes with an unsupervised physics-constrained training approach, presenting a step towards fast and accurate mesh simulations for PDEs without a requirement for pre-computed datasets. In particular, we will (a) present an efficient, finite-difference-based method to compute the directional derivatives for the physics-constrained losses on the mesh, (b) implement a physics-constrained unsupervised training method operating on meshes, and (c) quantitatively evaluate our framework on several PDE examples in 1D and 2D, namely the Heat, Eikonal, and Burger equations.


\section{Methods}
\label{sec:meth}

Let us first define the PDE problem we are solving. We use the following form for a PDE $f_{\mathrm{PDE}}$ and Boundary Conditions (BC) $f_{\mathrm{BC}}$:
\begin{equation}
\begin{aligned}
    f_{PDE} \left(t, x, u, \frac{\partial u}{\partial t}, \frac{\partial u}{\partial x}, \frac{\partial^2 u}{\partial x^2}, ..., \frac{\partial^n u}{\partial x^n} \right) &= 0
    \\
    f_{BC} \left(t, x, u \frac{\partial u}{\partial t}, \frac{\partial u}{\partial x}, \frac{\partial^2 u}{\partial x^2}, ..., \frac{\partial^n u}{\partial x^n} \right) &= 0
\end{aligned}
\label{eq:pde}
\end{equation}

Based on this formulation, we can define the physics-constrained loss similarly to previous work \citep{raissi2019physics, wandel2020learning} with tunable parameters $\alpha$ and $\beta$:
\begin{equation}
\begin{aligned}
    \mathcal{L} &= \alpha \left\| f_{\mathrm{PDE}} \left(t, x, u, \frac{\partial u}{\partial t}, \frac{\partial u}{\partial x}, \frac{\partial^2 u}{\partial x^2}, ..., \frac{\partial^n u}{\partial x^n} \right) \right\|^2
    \\
    &+ \beta \left\| f_{\mathrm{BC}} \left(t, x, u, \frac{\partial u}{\partial t}, \frac{\partial u}{\partial x}, \frac{\partial^2 u}{\partial x^2}, ..., \frac{\partial^n u}{\partial x^n} \right) \right\|^2 \quad .
\end{aligned}
\label{eq:pdeloss}
\end{equation}
For the sake of simplicity we use in \cref{eq:pde,eq:pdeloss} a 1-dimensional notation for space $x$ and variable $u$, but the notation can be trivially extended to higher dimensions. We will refer to them in the following as $\vec{x}$ and $\vec{u}$.

We chose to use the Graph Network architecture from \citet{sanchez2018graph, pfaff2021learning} due to its inductive bias conducive to mesh learning. A solution of the PDE is represented as values $\vec{u}_t$ on a discretized mesh at a timestep $t$. This discretized mesh additionally has a position $\vec{x}$ and extra vertex information $\vec{m}$. Vertex information in our case is what type the node is, \emph{e.g.}, a normal node, an enclosing wall, an inflow, an outflow, or an internal obstacle. We store this nodal information as vertex attributes $V$ of graph $M^t = (V, E)$ representing this mesh at timestep $t$. Edge attributes are represented by $E$, which in our case contain the edges of the mesh and relative mesh position between nodes in the mesh.

We model the time-dependent PDE problem from \cref{eq:pde} with a neural network $g_{\mathrm{NN}}(\vec{x}, \vec{u}_t, \vec{m})$, predicting a residual $\Delta \vec{u}$ which computes the solution at the next timestep $\vec{u}_{t+1}=\vec{u}_t+\Delta \vec{u}$ given the input graph $M^t$. These two values, $\vec{u}_{t}$ and $\vec{u}_{t+1}$, are then used to compute the physics-constrained loss from \cref{eq:pdeloss}. This computation requires an appropriate approximation for the derivatives. 

In our formulation in \cref{eq:pde} we chose the limitation to just include first-order time derivatives, as in this work we work with solutions $\vec{u}_{t}, \vec{u}_{t+1}$, which can only discretize a first-order time derivative in a satisfying manner. This simplification can, however, be easily extended by predicting more than one timestep $\vec{u}_t, ..., \vec{u}_{t+k}$ and using these multiple predictions for higher-order derivatives (up to $k$-th order derivatives). By only using the solutions at timestep $\vec{u}_{t}$ and the next timestep $\vec{u}_{t+1}$, we can discretize the equations with the unconditionally stable implicit/backward Euler method \citep{butcher2016numerical}, resulting in discretized functions $\hat{f}_{\mathrm{PDE}} (\vec{u}_{t}, \vec{u}_{t+1})$ and $\hat{f}_{\mathrm{BC}} (\vec{u}_{t}, \vec{u}_{t+1})$. This option alleviates any issues that may occur when choosing inappropriate temporal $\Delta t$ or spatial $\Delta x$ discretization. Solving time-independent systems in our formulation results in setting $\Delta t \to \infty$ so it solves for the steady-state solution (independent of the initial condition). In the next \cref{sec:spatialgrad} we will define an approach of how to compute the derivatives for our PDE in our graph representation.

Once we have created the framework to compute for a discretized PDE and BC losses on the graph, we will adapt the training algorithm from \citet{wandel2020learning, wandel2021spline} in \cref{sec:unsuptrain} to evaluate our approach on simple PDE examples with fixed initial conditions in \cref{sec:exp}. An outline of how the randomized environment is used and how the network prediction feeds into the physics-constrained loss can be found in \cref{fig:outline}.


\subsection{Spatial Gradients in a Mesh}
\label{sec:spatialgrad}

\begin{figure}[tb]
\centering
\includegraphics[width=\columnwidth]{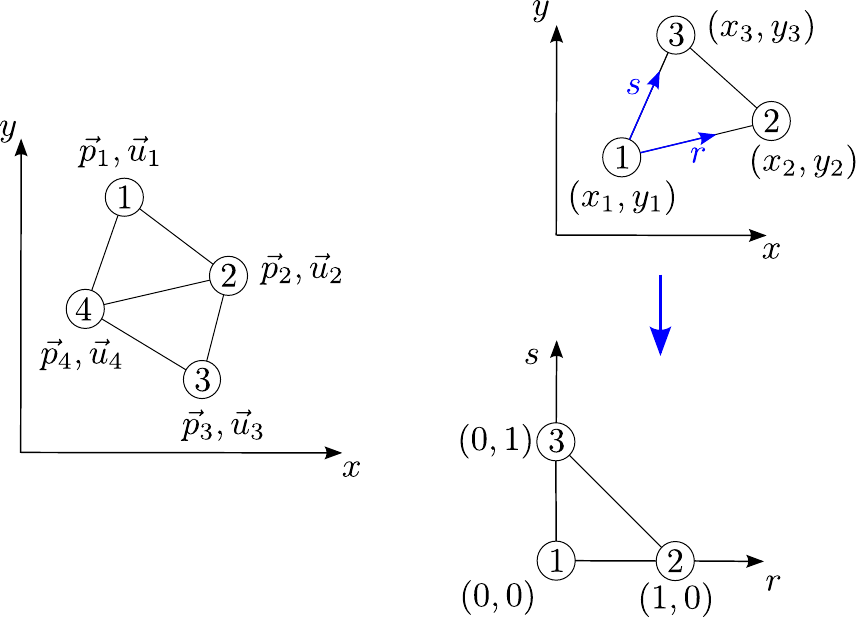}
\vspace{-1.5em}
\caption{Left: Every node stores the position $\vec{p}_i$ and value $\vec{u}_i$ and averages the computed finite difference over all neighbors $\mathcal{N}$. Right: Transformation of triangles from global coordinates ($x$,$y$) to local natural coordinates ($r$,$s$). We leverage the fact that the finite difference derivative is well-defined within the local coordinates.}
\label{fig:graphgrad}
\end{figure}

We predict values for the whole mesh (not just individual points), taking the topology and environment description into account. This choice will allow for far better generalizability than PINNs, however, if we use autodifferentiability to compute gradients, as in \citet{raissi2019physics}, backward passes will be far more complex, as gradients are determined by all nodes in the mesh, which makes the practical usage of autodifferentiability on meshes infeasible. Rather than predicting a solution from position/time inputs and computing derivatives through autodifferentiation, a more simple but also limiting approach is to use finite differences\,\citep{wandel2020learning}. First, we define the operation of a finite spatial difference on a mesh as in \cref{fig:graphgrad}-Left. We exemplify our approach in 2D, but it can be similarly extended to any dimension (\cref{app:grad3D} shows the extension to 3D). This way we can compute all PDE terms using finite differences.

A naive approach for a mesh with nodal values $\vec{u}_i$ and positions $\vec{x}$ (see \cref{fig:graphgrad}-Left) would have spatial derivatives of value $\vec{u}$ at node $i$ described as follows:
\begin{equation}
\begin{aligned}
    \frac{\partial \vec{u}_i}{\partial x} &\approx \frac{1}{|\mathcal{N}|} \sum_{j \in \mathcal{N}} \frac{\left( \vec{u}_j - \vec{u}_i \right)_x}{\left( \vec{p}_j - \vec{p}_i \right)_x}
    \\
    \frac{\partial \vec{u}_i}{\partial y} &\approx \frac{1}{|\mathcal{N}|} \sum_{j \in \mathcal{N}} \frac{\left( \vec{u}_j - \vec{u}_i \right)_y}{\left( \vec{p}_j - \vec{p}_i \right)_y}
\end{aligned}
\label{eq:fdfail}
\end{equation}
However, this approach quickly runs into an issue when two nodes align on either x- or y-axis, as the distance $\Delta \vec{p}$ on the axis is then approaching zero, causing some values of the finite difference to explode. Instead, we rely on the well-understood theory of finite elements \citep{ern2004theory}, and apply it on our triangular mesh. The core idea that we will adopt is to transform the triangular elements from their global coordinates $x, y$ into natural coordinates $r, s$, where we define the finite difference operation as if it were on a regular grid (see \cref{fig:graphgrad}-Right). A shape function is used to interpolate values within the finite element. We illustrate our examples with linear triangular elements. The extension to second-order elements would allow for second-order derivatives, however, it would introduce extra nodes in the mesh, increasing the computational complexity significantly.

In the case of linear triangular elements in 2D (extended to 1D in \cref{app:grad1D} and 3D in \cref{app:grad3D}), we transform triangles from their global $x, y$ coordinates to natural local $r, s$ coordinates. We will summarize the approach in the following and a full explanation can be found in \cref{app:gradder}. The triangles with values $u_i$ at the vertices can be described as a function of their local coordinates as follows:
\begin{align}
    \begin{pmatrix} x(r,s) \\ y(r,s) \end{pmatrix} &= \begin{pmatrix} x_1 + (x_2 - x_1) r + (x_3 - x_1) s \\ y_1 + (y_2 - y_1) r + (y_3 - y_1) s \end{pmatrix}
    \\[0.5em]
    u(r,s) &=  u_1 + (u_2 - u_1) r + (u_3 - u_1) s
\label{eq:localtransform}
\end{align}
Given a value $u$ that varies throughout the finite element, we can write its total derivative with the chain rule:
\begin{align*}
    \begin{pmatrix} \frac{\partial u}{\partial r} \\ \frac{\partial u}{\partial s} \end{pmatrix}  
    &= \begin{pmatrix} 
        \frac{\partial u}{\partial x} \frac{\partial x}{\partial r} + \frac{\partial u}{\partial y} \frac{\partial y}{\partial r} \\ 
        \frac{\partial u}{\partial x} \frac{\partial x}{\partial s} + \frac{\partial u}{\partial y} \frac{\partial y}{\partial s} 
    \end{pmatrix}  
    = \begin{pmatrix}
        \frac{\partial x}{\partial r} & \frac{\partial y}{\partial r} \\
        \frac{\partial x}{\partial s} & \frac{\partial y}{\partial s}
    \end{pmatrix}
    \begin{pmatrix} \frac{\partial u}{\partial x} \\ \frac{\partial u}{\partial y} \end{pmatrix} 
    \\
    &= \begin{pmatrix}
        x_2 - x_1 & y_2 - y_1 \\
        x_3 - x_1 & y_3 - y_1
    \end{pmatrix}
    \begin{pmatrix} \frac{\partial u}{\partial x} \\ \frac{\partial u}{\partial y} \end{pmatrix} 
    =: \mathbf{J} \begin{pmatrix} \frac{\partial u}{\partial x} \\ \frac{\partial u}{\partial y} \end{pmatrix} 
\end{align*}
We can compute $\mathbf{J}$ using the known positions of the nodes. Our initial objective is to calculate the derivatives $\frac{\partial u}{\partial x}$ and $\frac{\partial u}{\partial y}$ w.r.t. the global coordinates $x$ and $y$:
\begin{equation}
\begin{aligned}
    \nabla u &= \begin{pmatrix} \frac{\partial u}{\partial x} \\ \frac{\partial u}{\partial y} \end{pmatrix}  = \mathbf{J}^{-1} \begin{pmatrix} \frac{\partial u}{\partial r} \\ \frac{\partial u}{\partial s} \end{pmatrix} 
    \\
    &= \begin{pmatrix}
        x_2 - x_1 & y_2 - y_1 \\
        x_3 - x_1 & y_3 - y_1
    \end{pmatrix}^{-1} \begin{pmatrix} u_2 - u_1 \\ u_3 - u_1 \end{pmatrix} 
\end{aligned}
\label{eq:spatgrad}
\end{equation}
The matrix $\mathbf{J}$ is always invertible as long as the area of the triangular element is nonzero (and therefore $\det{\mathbf{J}} \neq 0)$. 

This gradient $\nabla u$ is constant throughout the triangular element. For the gradient values at each node of the mesh, we will average the gradients of all neighboring elements. The detailed averaging procedure can be found in \cref{app:gradder}. In 1D, this approach is exactly equivalent to the central finite difference method, a second-order approximation. Similar to the previous approach for computing the first-order gradient, we will compute the second derivative by setting $u$ equal to the first derivative.

\subsection{Physics-constrained Unsupervised Mesh Learning}
\label{sec:unsuptrain}

We compute the physics-constrained loss $f_{\mathrm{PDE}}$ specified by \cref{eq:pdeloss} using the previously defined directional graph gradients for the spatial derivative terms. The PDE is solved when the function $f_{\mathrm{PDE}}$ reaches 0, hence $f_{\mathrm{PDE}}$ indirectly also gives us a measure of convergence for the solution. But most PDEs require not only their equation to hold, but additionally have Boundary Conditions (BC) to fulfill. Following previous approaches~\cite{raissi2019physics, wandel2020learning, pmlr-v144-pizzuto21a}, we introduce the BC as an additional loss term $f_{\mathrm{BC}}$ to be minimized. This BC can be both Dirichlet or Neumann BC. The total BC loss will then be a linear combination of the different BC terms. Each BC loss term has its own weight. Concrete examples can be seen in \cref{sec:exp}.

Having built the total loss $\mathcal{L}$ as in \cref{eq:pdeloss}, we now tackle the problem of how the graph network is trained. As our network is designed to perform discrete timestepping, it can predict a sequence of solutions by itself given some initial condition. Because the network should be able to generalize to a variety of initial conditions, we randomly sample these initial conditions, and create a dataset of solutions $\vec{u}_k$ with nodal information $\vec{m}_k$ at $t = 0$. Next, we sample a minibatch from this generated dataset, predict one step forward in time using our network, and add the result back into the dataset. This randomized environment as a dataset is growing over time and assures that the existing predictions in the dataset will become more and more accurate. \citet{wandel2020learning} has used this approach on a regular grid to create a generalizable fluid solver within a limited range of fluid parameters.

\begin{algorithm}[ht]
\setstretch{1.05}
   \caption{Physics-constrained Unsupervised Training}
   \label{alg:train}
\begin{algorithmic}
   \STATE {\bfseries Inputs:} 
   \STATE \quad Mesh positions $\vec{x}$
   \STATE \quad Neural network $g_{\mathrm{NN}}(\vec{x}, \vec{u}_{\;t}, \vec{m})$
   \STATE \quad Number of randomized domains $N_k$
   \STATE \quad Number of timesteps $N_t$
   \STATE   
   \STATE Randomize initial conditions $\Omega^0 = \left\{ \vec{u}_k^{\;0}, \vec{m}_k^0 \right\}_{k \in \{1...N_k\}}$
   \STATE Set environment of all timesteps to initial condition \hspace{1em} $\Omega^t = \left\{ \vec{u}_k^{\;t}, \vec{m}_k^t \right\}_{k \in \{1...N_k\}} := \Omega^0 \hspace{0.5em} \forall t \in {1 ... N_t}$
   \STATE Create randomized environments $\Omega = \left\{ \Omega^t \right\}_{t \in \{1...N_t\}}$
   \REPEAT
   \STATE Sample batch $\left\{\Omega_{b}^t\right\}_{b \subset{\{1...N_k\}}}$ from randomized environments $\Omega$ at fixed time $t$
   \STATE
   \STATE Batch loss $\mathcal{L}_{total}$ = 0
   \FOR{{\bfseries all} $\left\{ \vec{u}_b^{\;t}, \vec{m}_b^t \right\}$ {\bfseries in} $\Omega_b$}
   \STATE Predict $\vec{u}_b^{\;t+1} = g_{\mathrm{NN}}(\vec{x}, \vec{u}_b^{\;t}, \vec{m}_b^t)$
   \STATE $\mathcal{L}_{total} \mathrel{{+}{=}} \mathcal{L}(\vec{u}_b^{\;t}, \vec{u}_b^{\;t+1})$
   \STATE Replace $\Omega^{t+1}$ with $\left\{ \vec{u}_b^{\;t+1}, \vec{m}_b^t \right\}$
   \ENDFOR
   \STATE
   \STATE Optimize $g_{\mathrm{NN}}$ based on $\mathcal{L}_{total}$ 
   
    \UNTIL{maximum epochs reached}
\end{algorithmic}
\end{algorithm}

We described above the training algorithm based on \citet{wandel2020learning}. In our experiments, however, we noticed that this training scheme does not always perform well. When we test our graph network with directional gradients, we "overfit" the network on a single initial condition. Consider we have a time domain of interest ranging between $[0, 1]$ with $\Delta t = 0.01$. On this interval, we have a completely non-uniform sampling of solutions, where solutions near the start of the simulation $t = 0$ are sampled much more frequently than near the end $t = 1$. In our case of a single initial condition, it made training the graph network nearly impossible. Hence we adjusted the implementation to consider a fixed time interval $[t_a, t_b]$, filling our initial dataset with $u(t, x) = u(t_a, x) \hspace{0.5em} \forall t \in [t_a, t_b]$. Note that this time interval limitation does not restrict the maximum run length at test-time. The detailed procedure of our training scheme can be seen in \cref{alg:train}.

\section{Experimental Setup}
\label{sec:exp}

To verify our physics-constrained unsupervised learning  approach, we will evaluate our graph network using our training scheme on several PDE problems that are described on a mesh. We first describe the implementation of our architecture and the specific training setup, and then follow with the example PDE objectives used for the experimental validation.

\subsection{Network Architecture and Training}

The graph network we use has a slight modification, as we noticed spiking behavior when using the vanilla graph network \citep{sanchez2018graph}. This behavior is due to the aggregation during message passing being a summation, and our triangular mesh not having uniform node valence. Hence some nodes have more neighbors than others. We change the aggregation to an averaging operation, which makes sure that the network predictions are much smoother. An alternative we considered, before changing the aggregation function, was to have a "warm start" for the network and have it "train" for 1000 epochs to have a uniform constant zero output. 

Our networks are defined with 4 MLP layers each in encoder, processor, and decoder. Latent dimension is 16, and we use a learning rate of $10^{-3}$ with Adam optimizer. The implementation is in PyTorch. By default we use a spatial discretization $\Delta x = 0.05$ and a temporal discretization $\Delta t = 0.01$. For now our meshing algorithms create a regular triangular mesh on the domain. To compare with a numerical baseline, we use the Python API of FEniCS Project PDE solver \citep{alnaes2015fenics}. We solve the PDE on a high resolution mesh in FEniCS, upscaled 10x in every dimension, and then downsample to compute Mean Squared Errors (MSE) compared with our network predictions.

\subsection{Example PDE Objectives}

Next, we will present some of the PDEs that will be trained using this framework. We start out by solving single, specific initial conditions that we can benchmark against the FEniCS baseline. We then design a PDE objective with randomized initial conditions, and validate its generalization performance on a specific scenario.

\paragraph{Heat Equation in 1D} A simple time-dependent PDE to test the second derivative: 
\begin{equation}
\begin{aligned}
    \frac{\partial u}{\partial t} - 0.4 \frac{\partial^2 u}{\partial x^2} &= 0
    \\[0.5em]
    u(t=0, x) &= \frac{sin(2\pi x)}{2\pi x}
\end{aligned}
\end{equation}
The domains for the time and space variables are defined as  $t \in [0, 1]$ and $x \in [-1, 1]$. The temporal BC above translates into an initial condition for the domain. Following \cref{eq:pdeloss} and given that we do not have any spatial BCs:
\begin{equation}
\begin{aligned}
    \mathcal{L}_{Heat} = \;
    &\alpha \left\| \frac{\partial u}{\partial t} - 0.04 \frac{\partial^2 u}{\partial x^2} \right\|^2 
\end{aligned}
\label{eq:heat1Dloss}
\end{equation}
We choose $\alpha=1$ in most cases, especially since no BC terms are present in this objective.

\paragraph{Eikonal Equation in 1D} This PDE is often seen in wave propagation, and can be interpreted as solving  shortest-path problems.
\begin{equation}
\begin{aligned}
    \frac{\partial u}{\partial t} + \left| \frac{\partial u}{\partial x} \right| - 1 &= 0
    \\[0.5em]
    u(t=0, x) &= 0
    \\[0.5em]
    u(t, x=-1) = u(t, x=1) &= 0
\end{aligned}
\end{equation}
The domains for the time and space variables are defined as $t \in [0, 1]$ and $x \in [-1, 1]$. We now consider our spatial boundary loss as well:
\begin{equation}
\begin{aligned}
    \mathcal{L}_{Eikonal} = \;
    &\alpha \left\| \frac{\partial u}{\partial t} +  \left| \frac{\partial u}{\partial x} \right| - 1 \right\|^2 
    \\
    + &\beta \left\| u(t, x=-1) \right\|^2 + \left\| u(t, x=1) \right\|^2
\end{aligned}
\label{eq:eikonalloss}
\end{equation}
We once again choose $\alpha=1$, but now we have to choose $\beta$ appropriately so the BCs are not violated in the final solution. In practice, we found $\beta=100$ to work well.

\paragraph{Burger's Equation in 1D} This equation is an often-used, fundamental PDE modeling dissipative systems. We will test a specific 1D form that appears in the review by \citet{blechschmidt2021ways}:
\begin{equation}
\begin{aligned}
    \frac{\partial u}{\partial t} + u \frac{\partial u}{\partial x} - \left( \frac{0.01}{\pi} \right) \frac{\partial^2 u}{\partial x^2} &= 0
    \\[0.5em]
    u(t=0, x) &= -sin(\pi x)
    \\[0.5em]
    u(t, x=-1) = u(t, x=1) &= 0
\end{aligned}
\end{equation}
The domains for the time and space variables are defined as $t \in [0, 1]$ and $x \in [-1, 1]$. We consider our temporal and spatial boundary loss separately, hence our total loss is written as:
\begin{equation}
\begin{aligned}
    \mathcal{L}_{Burger} = \;
    &\alpha \left\| \frac{\partial u}{\partial t} + u \frac{\partial u}{\partial x} - \left( \frac{0.01}{\pi} \right) \frac{\partial^2 u}{\partial x^2} \right\|^2 
    \\
    + &\beta \left\| u(t, x=-1) \right\|^2 + \left\| u(t, x=1) \right\|^2
\end{aligned}
\label{eq:burgerloss}
\end{equation}
Here we have $\alpha=1, \beta=10$.


\paragraph{Heat Equation in 2D} This approach will be tested using randomized initial conditions. We proceed as follows:  we generate 500 randomized environments, each environment has between 1 and 4 circular obstacles and between 1 and 4 Gaussian heat sources (uniformly random chosen amount). We uniformly sample center in $[-0.8, 0.8]$ and radius in $[0.1, 0.3]$ of the obstacles. Similarly we sample $a, b, \vec{c}$ from the ranges $[0, 5], [10,  50], [-0.8, 0.8]^2$ respectively for the Gaussian heat sources defined by: $\vec{u}_{0,i}= a \cdot \exp{(-0.5b \cdot \| \vec{x} - \vec{c} \|^2)}$. The governing equations for this system of variable initial condition is:
\begin{equation}
\begin{aligned}
    \frac{\partial u}{\partial t} - 0.5 \nabla^2 u &= 0
    \\
    u(t, \vec{x}=\partial B) &= 0
\end{aligned}
\end{equation}
The domains for the time and space variables are defined as $t \in [0, 1]$ and $\vec{x} \in [-1, 1] \times [-1, 1]$. Now we not only set the BC at the boundary of the spatial domain to be zero, but also at the boundary of our randomly generated obstacles.
\begin{equation}
\begin{aligned}
    \mathcal{L}_{Heat} = \;
    &\alpha \left\|  \frac{\partial u}{\partial t} - 0.5 \nabla^2 u \right\|^2 
    \\
    + &\beta \sum_{x_i \in \partial B} \left\| u(t, x=x_i) \right\|^2
\end{aligned}
\label{eq:heat2dloss}
\end{equation}
For the above equation we use $\alpha=1, \beta=100$.




\section{Results and Discussion}
\label{sec:disc}

We start by validating the graph gradient we introduced in \cref{sec:spatialgrad} by seeing if a graph network can learn to solve simple PDEs. We introduced our physics-constrained mesh learning framework in \cref{sec:meth}, in this section we discuss the results we observed.

\begin{figure}[!bt]
    \centering
    \begin{subfigure}{0.47\columnwidth}
        \centering
        \includegraphics[width=\columnwidth]{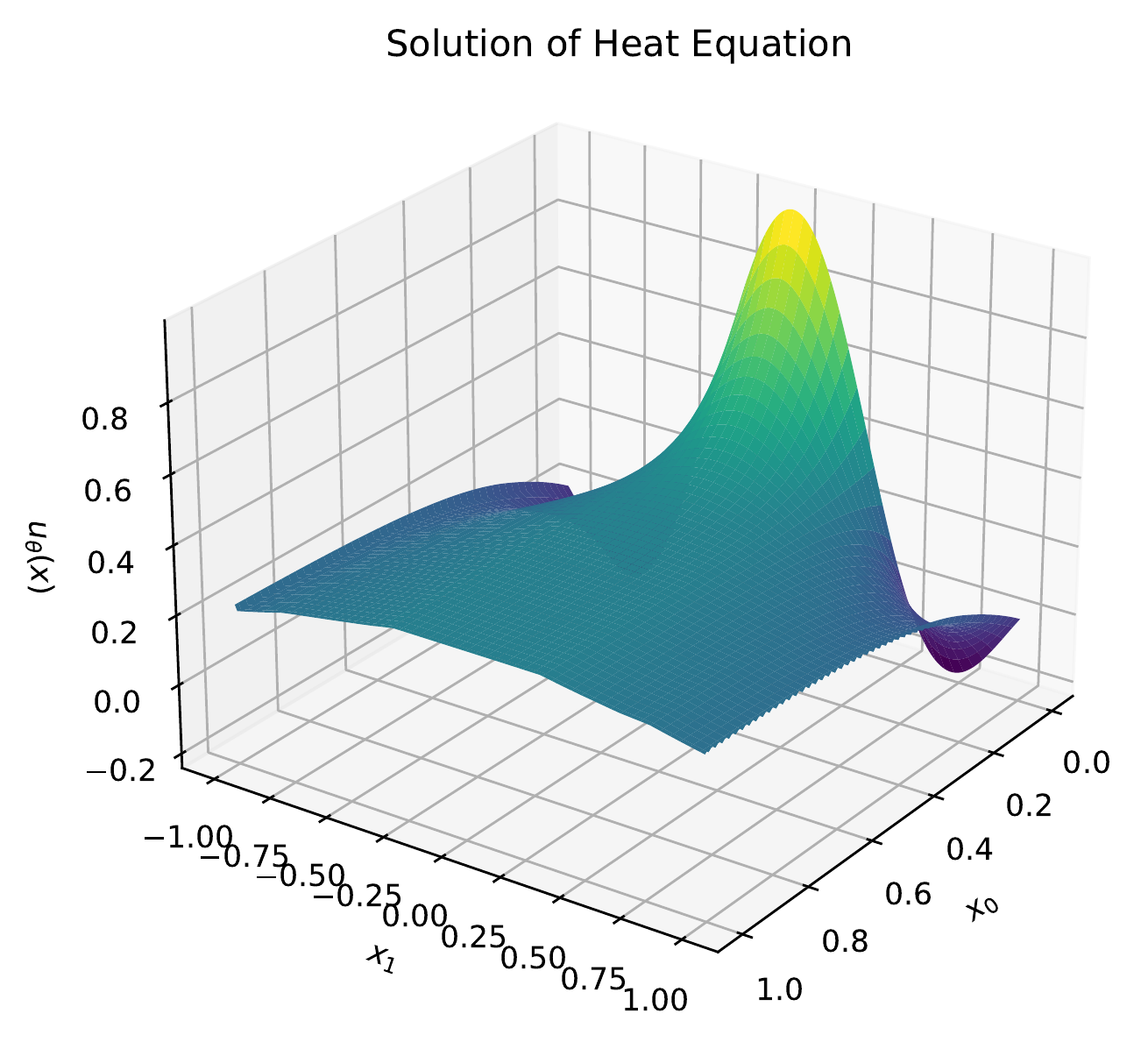}
    \end{subfigure}
    \begin{subfigure}{0.47\columnwidth}
        \centering
        \includegraphics[width=\columnwidth]{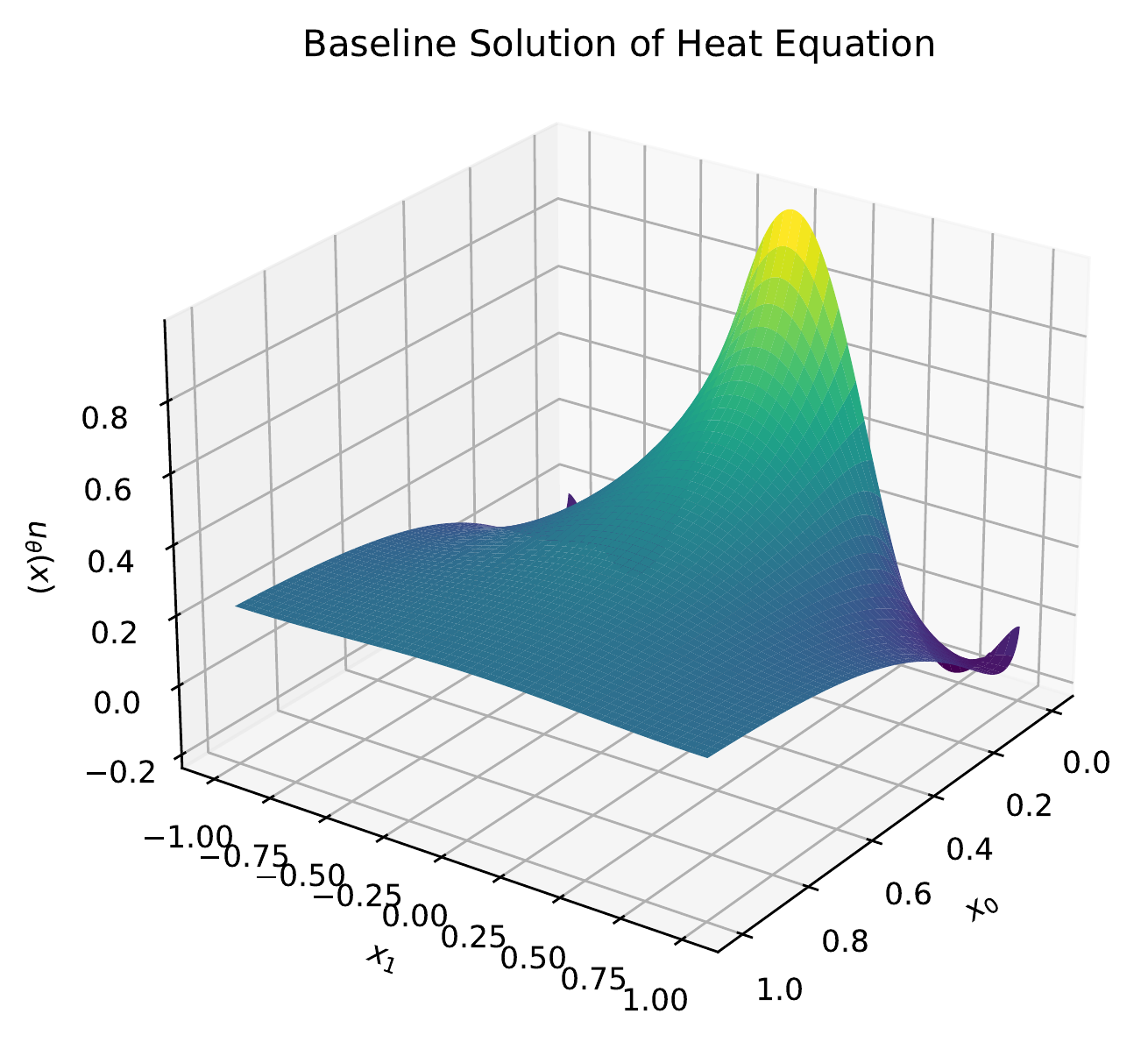}
    \end{subfigure}
    
    \caption{Solution to the Heat equation. Left: Prediction by the physics-constrained network. Right: Exact solution based on FEniCS for 1D heat equation.}
    \label{fig:heat1d}
\end{figure}

\begin{figure}[!tb]
    \centering
    \includegraphics[width=0.47\columnwidth]{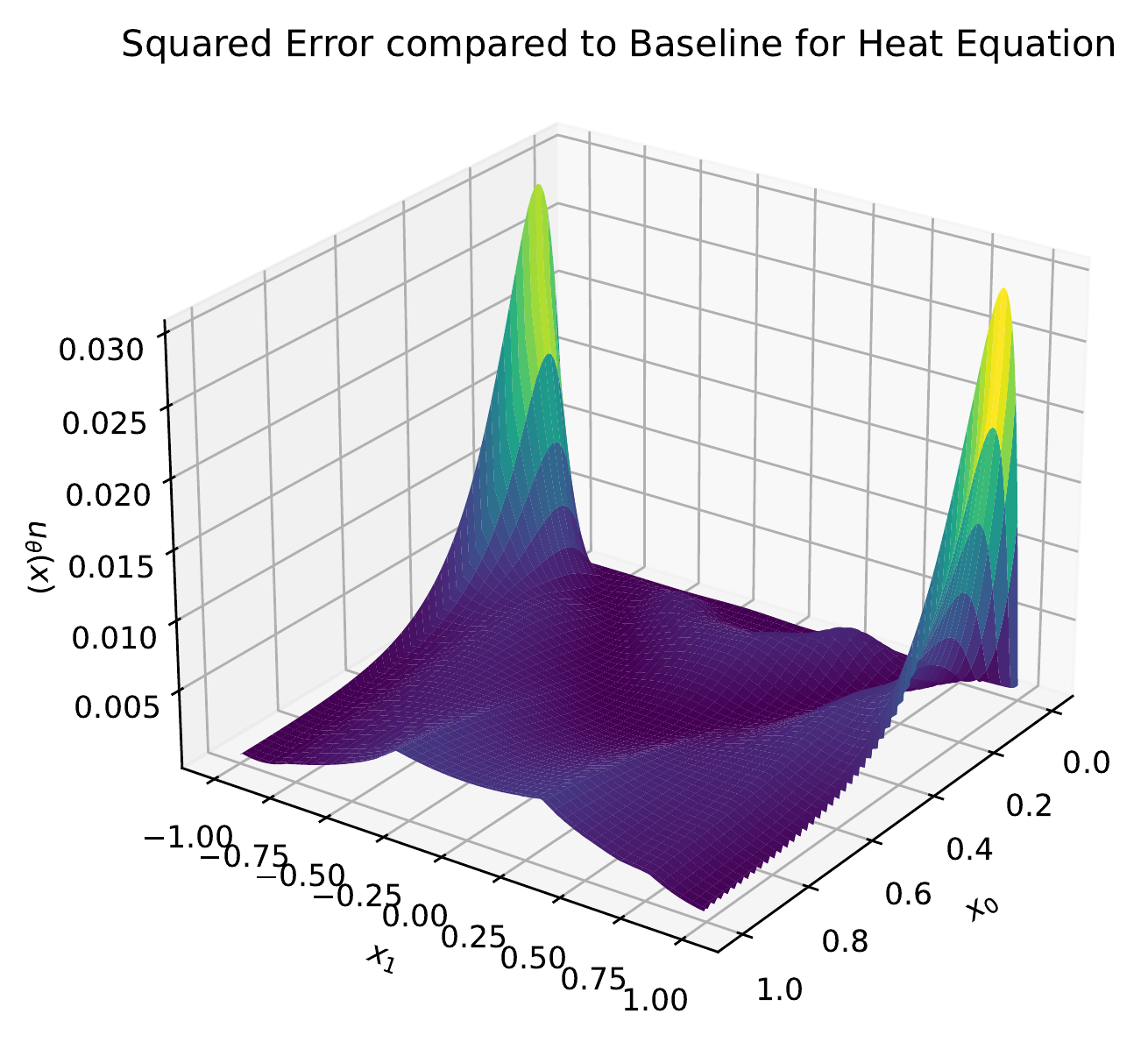}
    
    \caption{Squared error discrepancy of our method compared to baseline plotted at every vertex on the mesh for the Heat equation.}
    \label{fig:heat1derr}
\end{figure}

\paragraph{Heat Equation 1D} In \cref{fig:heat1d} we can see both the network prediction and the FEniCS baseline side by side. We show a surface plot of the solution $u$ over time, with $x_0$ being time and $x_1$ being space. Though the time-dependent behavior is correct, the error that is present as shown in \cref{fig:heat1derr} near the spatial boundary are problematic. We suspect that the finite difference based gradients are at fault, and the issue can be alleviated by extending our gradient computation to method using, \emph{e.g.} interpolation polynomials \citep{wandel2021spline}.

\begin{figure}[!tb]
    \centering
    \begin{subfigure}{0.47\columnwidth}
        \centering
        \includegraphics[width=\columnwidth]{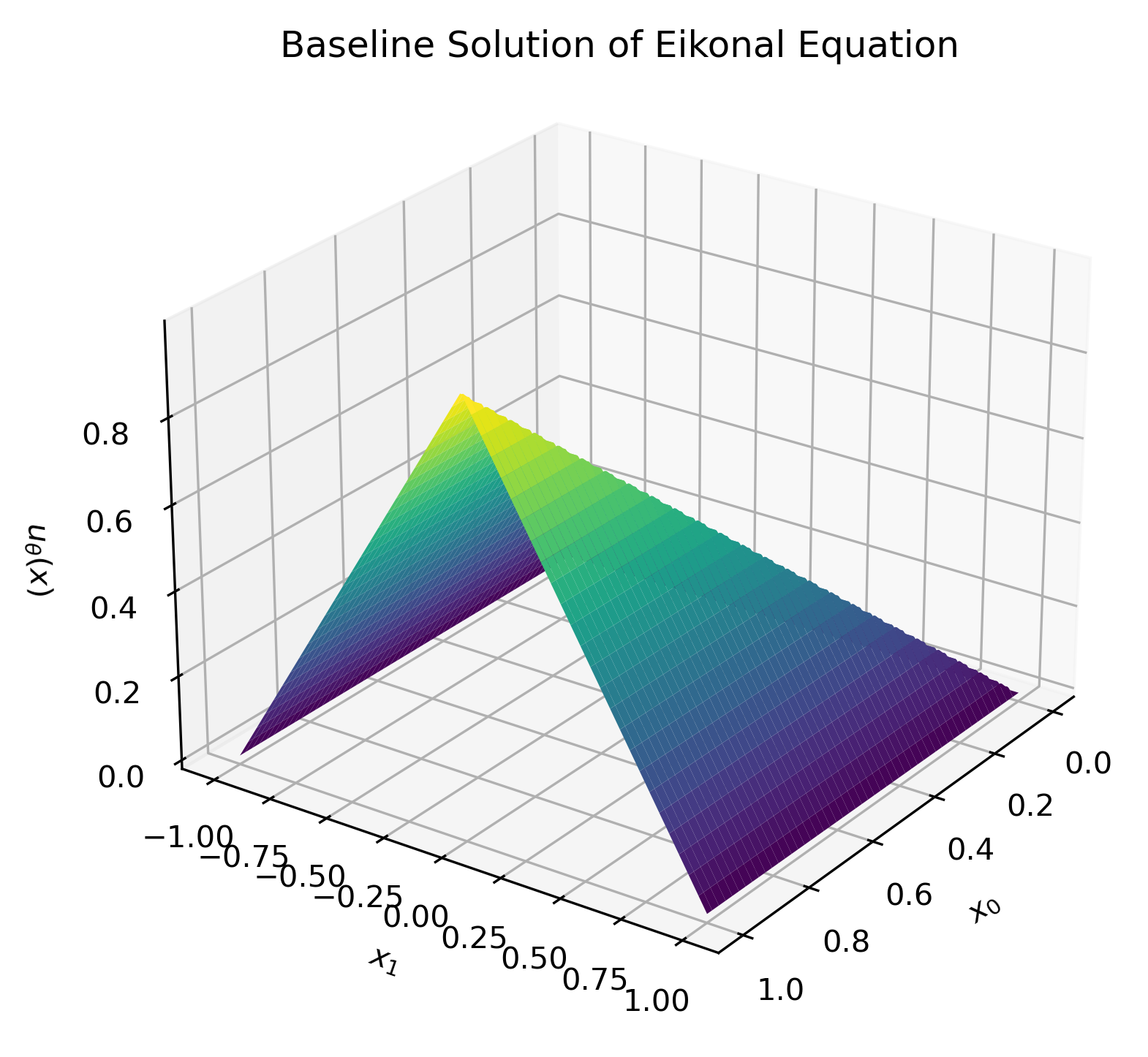}
    \end{subfigure}
    \begin{subfigure}{0.47\columnwidth}
        \centering
        \includegraphics[width=\columnwidth]{images/Eikonal1D_Baseline_PDE_Solution.png}
    \end{subfigure}
    
    \caption{Solution to the Eikonal equation. Left: Prediction by physics-constrained network. Right: Exact solution based on FEniCS for 1D Eikonal equation.}
    \label{fig:eik1d}
\end{figure}

\begin{figure}[!bt]
    \centering
    \includegraphics[width=0.47\columnwidth]{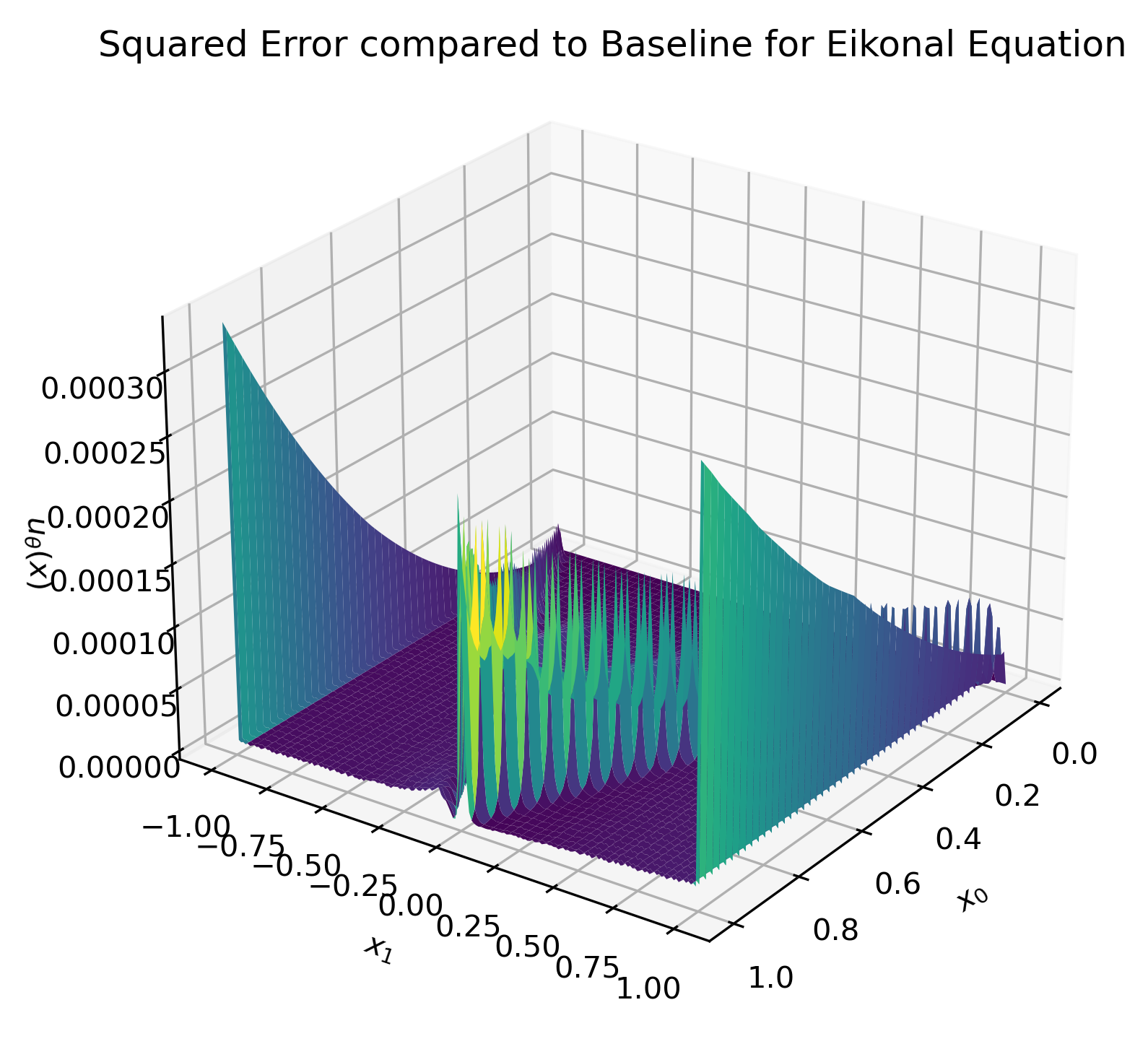}
    
    \caption{Squared error discrepancy of our method compared to baseline plotted at every vertex on the mesh for the Eikonal equation.}
    \label{fig:eik1derr}
\end{figure}

\paragraph{Eikonal Equation 1D} Similarly we validate the network on the Eikonal Equation. We see better results now in \cref{fig:eik1d} and \cref{fig:eik1derr}, where the main errors come from the discontinuity in the middle and the boundary conditions, both of which are expected given our finite difference gradient and BC optimization objective. The latter can be improved upon longer optimization.


\begin{figure}[!tb]
    \centering
    \begin{subfigure}{0.47\columnwidth}
        \centering
        \includegraphics[width=\columnwidth]{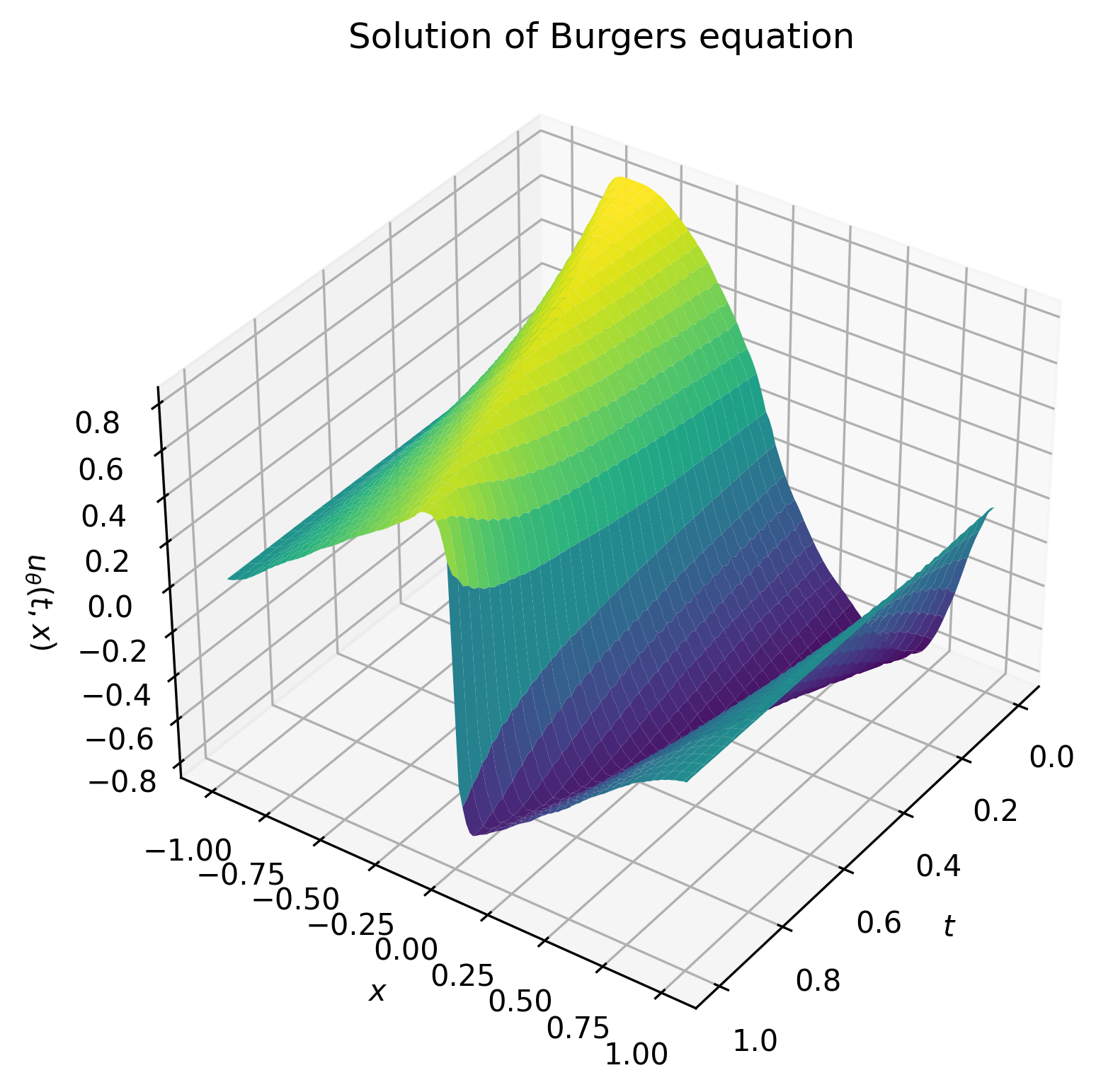}
    \end{subfigure}
    \begin{subfigure}{0.47\columnwidth}
        \centering
        \includegraphics[width=\columnwidth]{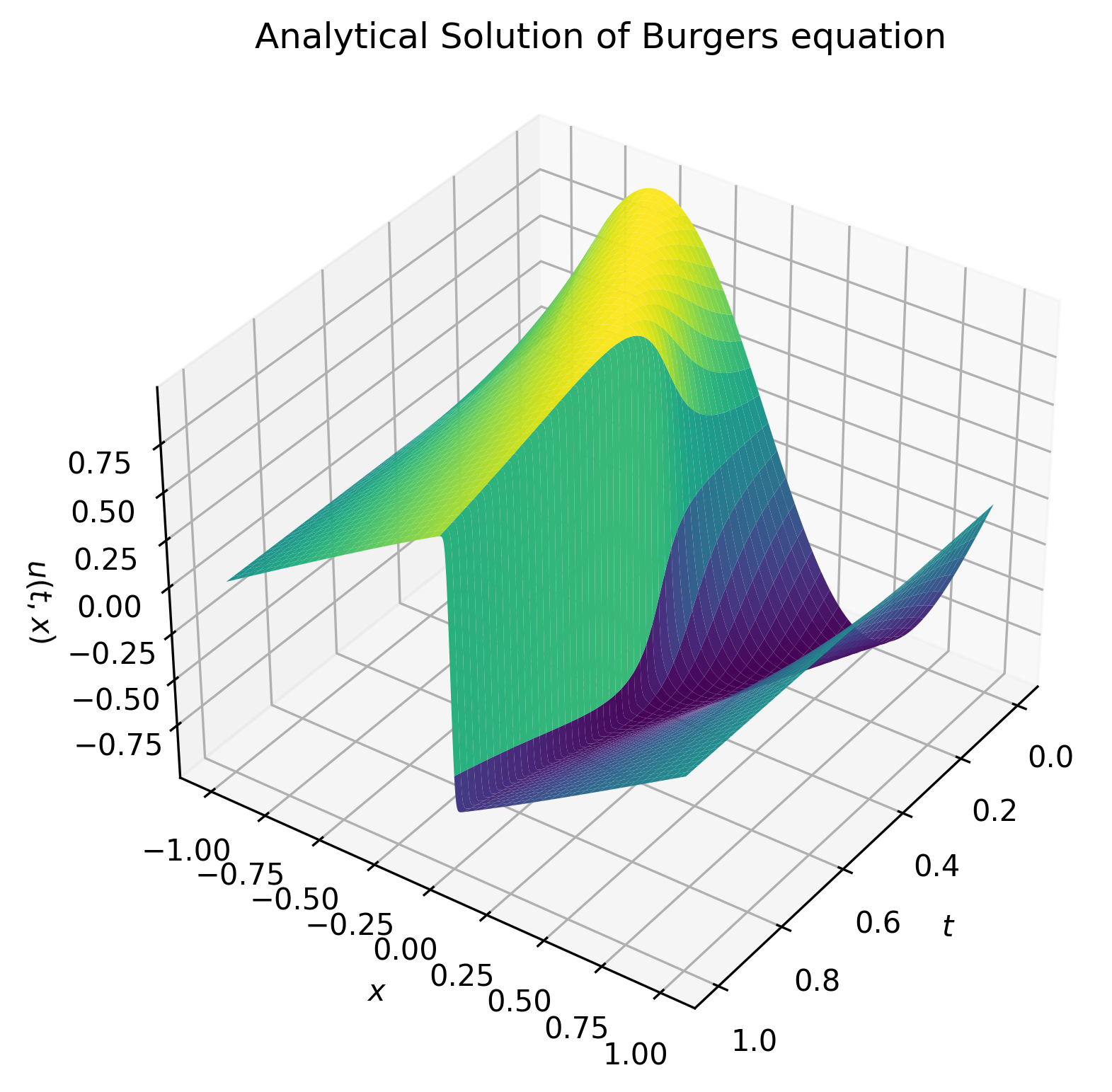}
    \end{subfigure}
    
    \caption{Solution to the Burgers equation. Left: Prediction by the physics-constrained network. Right: Exact solution using conventional solver FEniCS.}
    \label{fig:burger}
\end{figure}

\begin{figure}[!tb]
    \centering
    \includegraphics[width=0.47\columnwidth]{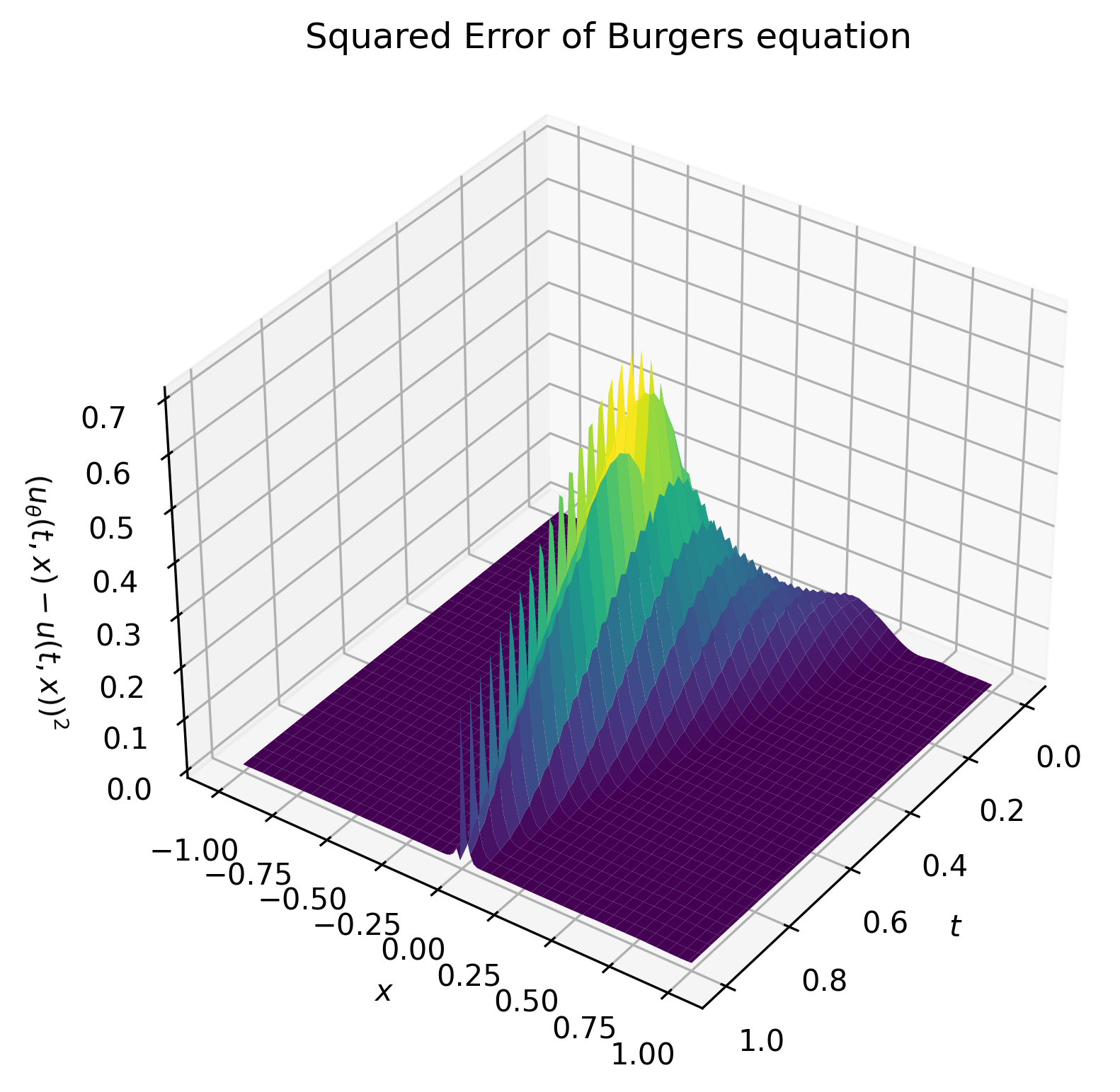}
    
    \caption{Squared error discrepancy of our method compared to baseline plotted at every vertex on the mesh for the Burgers equation.}
    \label{fig:burgerror}
\end{figure}

\paragraph{Burgers Equation} We solve for the previously defined loss for the Burgers Equation \cref{eq:burgerloss}. We are able to successfully create a qualitative similar solution to the baseline solution for the Burgers Equation (see Fig.~\ref{fig:burger}), though with significant error near the discontinuity (see Fig.~\ref{fig:burgerror}). This is to be expected, as the graph gradient acts as a finite difference, and near such drops the gradient approximation error will be high.

\begin{figure*}[!bt]
    \centering
    \begin{subfigure}{\textwidth}
        \centering
        \begin{subfigure}{\textwidth}
            \raisebox{0.0\textwidth}{\rotatebox[origin=t]{90}{\small{Initial Condition 1}}}
            \centering
            \begin{subfigure}{0.3\textwidth}
                \centering
                \begin{subfigure}[b]{0.55\textwidth}
                    \centering
                    \includegraphics[width=\textwidth]{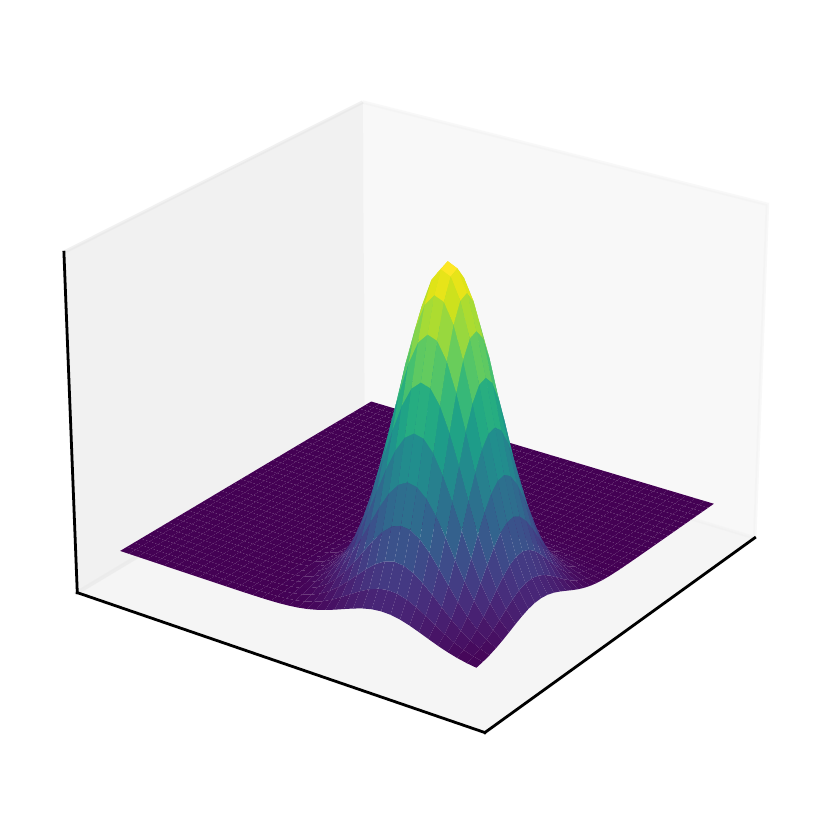}
                \end{subfigure}
                \begin{subfigure}[b]{0.4\textwidth}
                    \centering
                    \includegraphics[width=\textwidth]{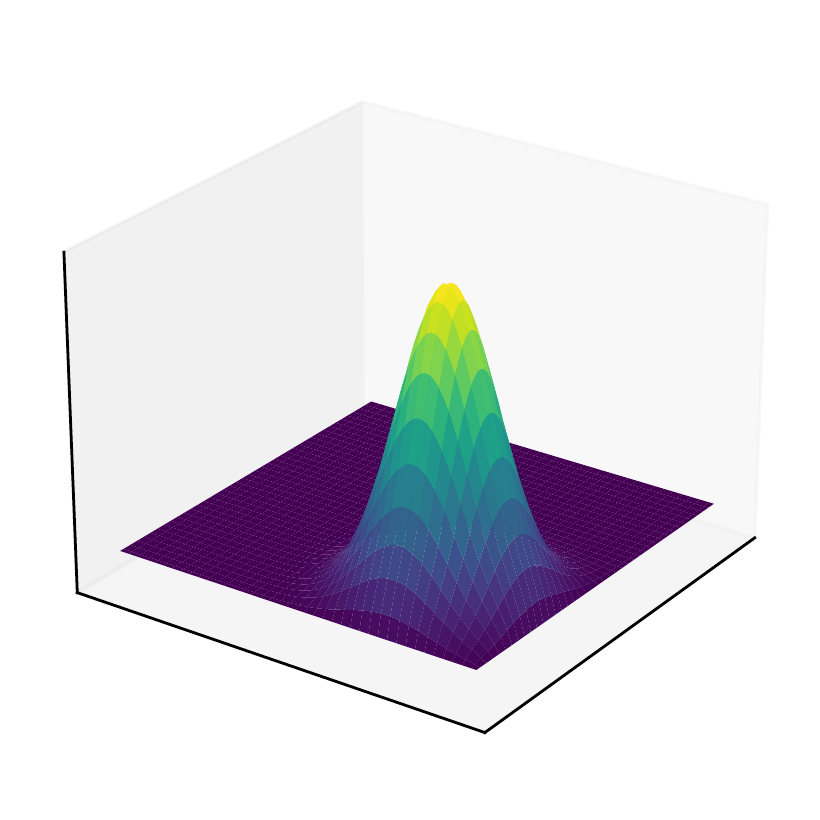}
                \end{subfigure}
            \end{subfigure}
            \begin{subfigure}{0.3\textwidth}
                \centering
                \begin{subfigure}[b]{0.55\textwidth}
                    \centering
                    \includegraphics[width=\textwidth]{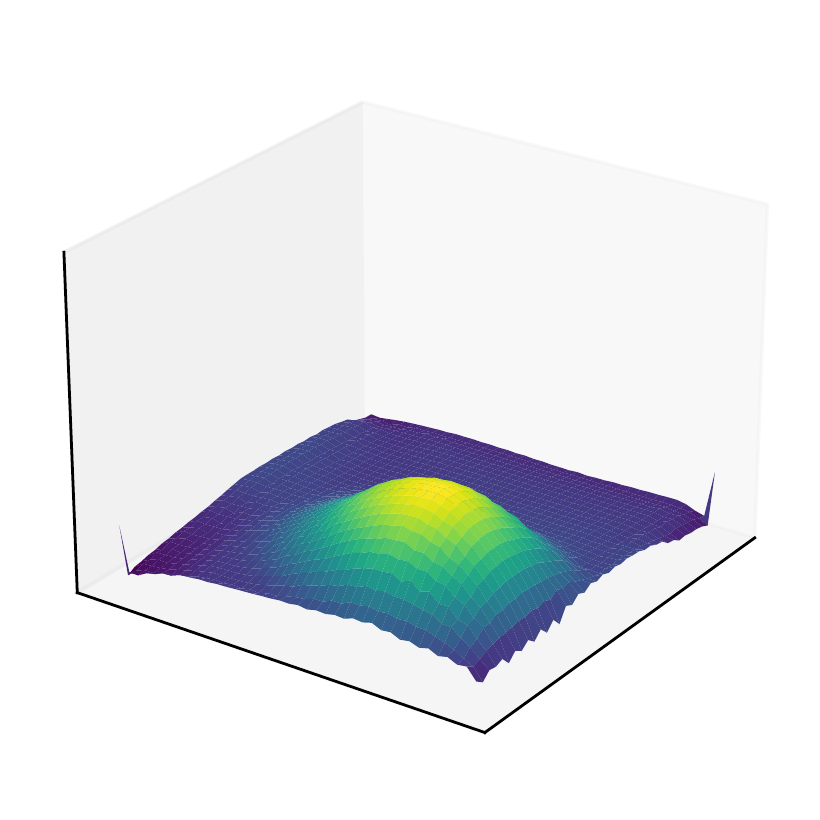}
                \end{subfigure}
                \begin{subfigure}[b]{0.4\textwidth}
                    \centering
                    \includegraphics[width=\textwidth]{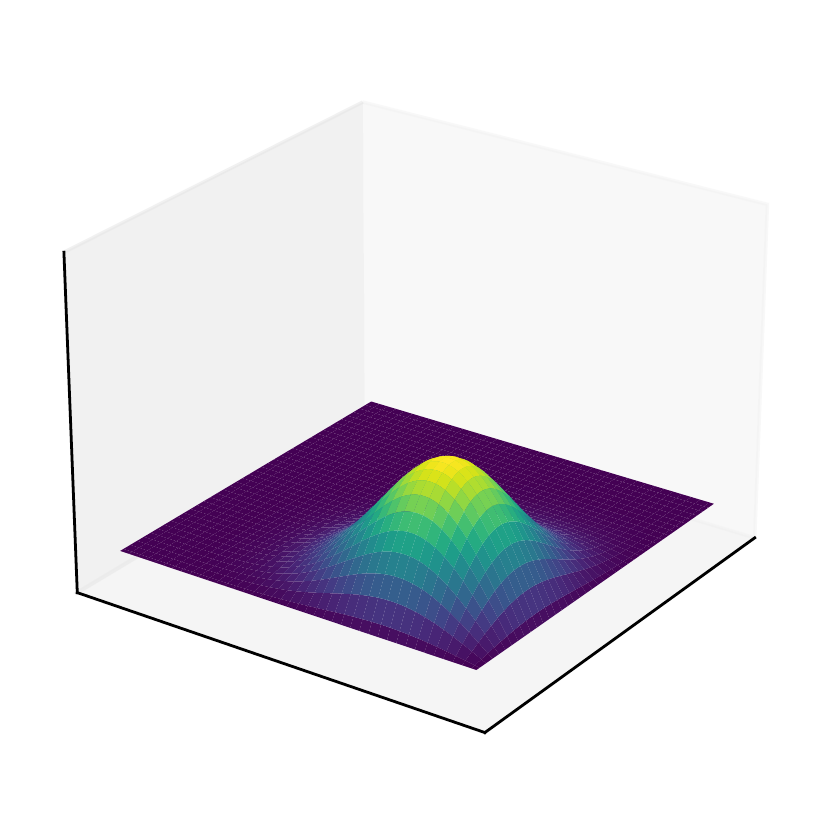}
                \end{subfigure}
            \end{subfigure}
            \begin{subfigure}{0.3\textwidth}
                \centering
                \begin{subfigure}[b]{0.55\textwidth}
                    \centering
                    \includegraphics[width=\textwidth]{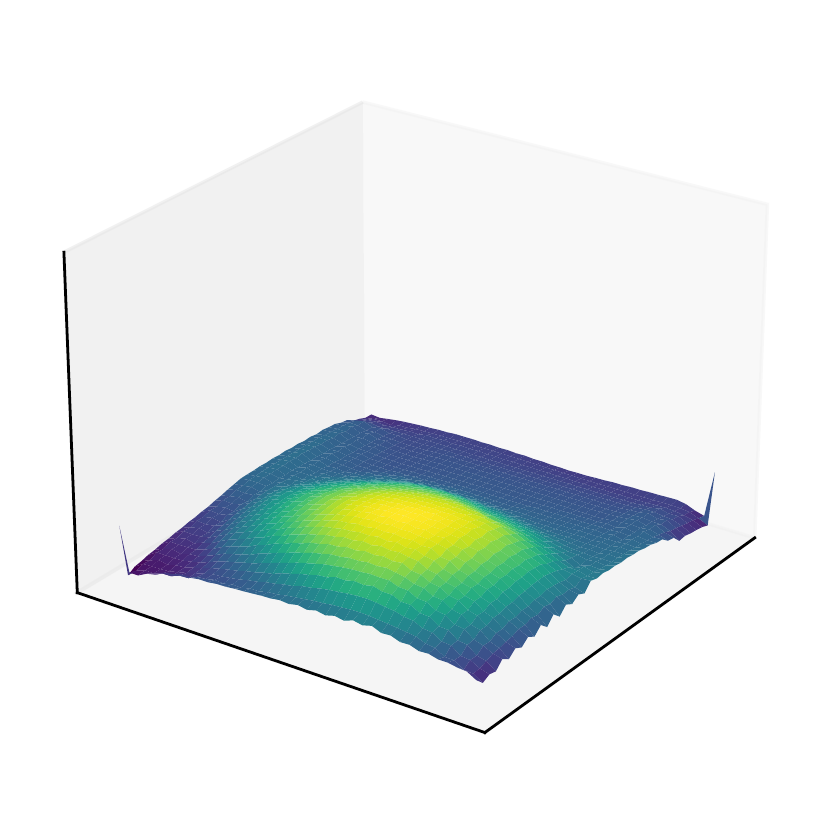}
                \end{subfigure}
                \begin{subfigure}[b]{0.4\textwidth}
                    \centering
                    \includegraphics[width=\textwidth]{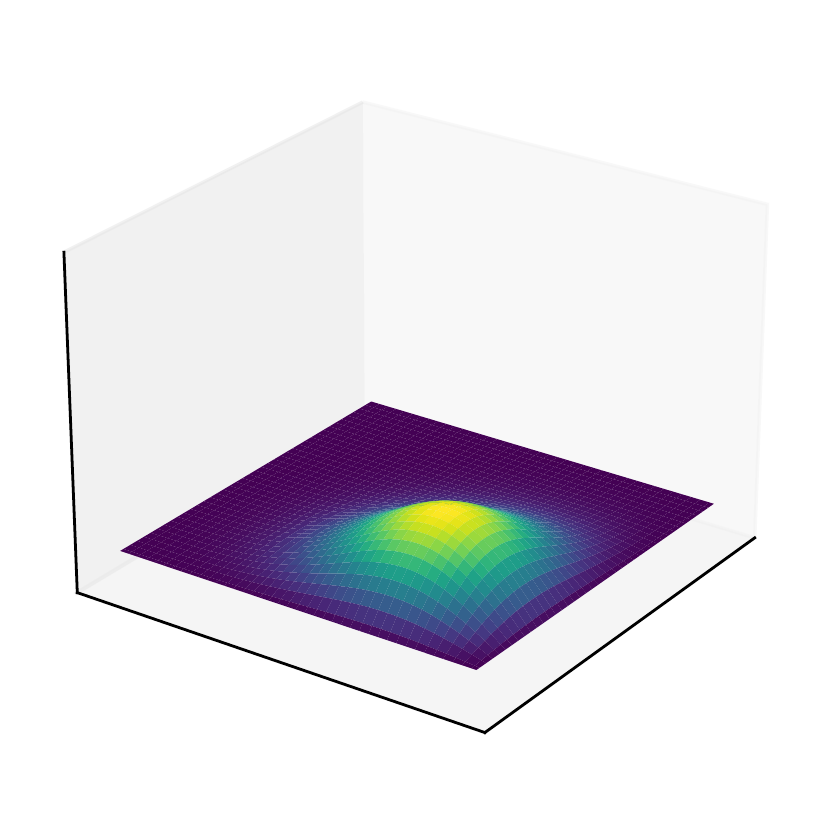}
                \end{subfigure}
            \end{subfigure}
        \end{subfigure}
    \end{subfigure}
    \\
    \begin{subfigure}{\textwidth}
        \centering
        \begin{subfigure}{\textwidth}
            \raisebox{0.0\textwidth}{\rotatebox[origin=t]{90}{\small{Initial Condition 2}}}
            \centering
            \begin{subfigure}{0.3\textwidth}
                \centering
                \begin{subfigure}[b]{0.55\textwidth}
                    \centering
                    \includegraphics[width=\textwidth]{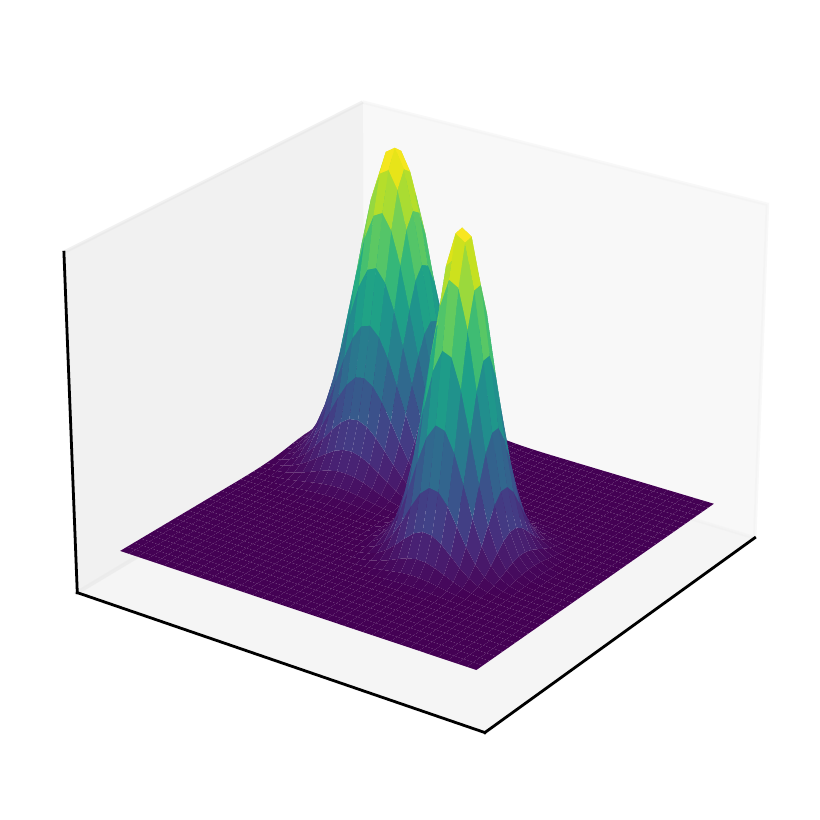}
                \end{subfigure}
                \begin{subfigure}[b]{0.4\textwidth}
                    \centering
                    \includegraphics[width=\textwidth]{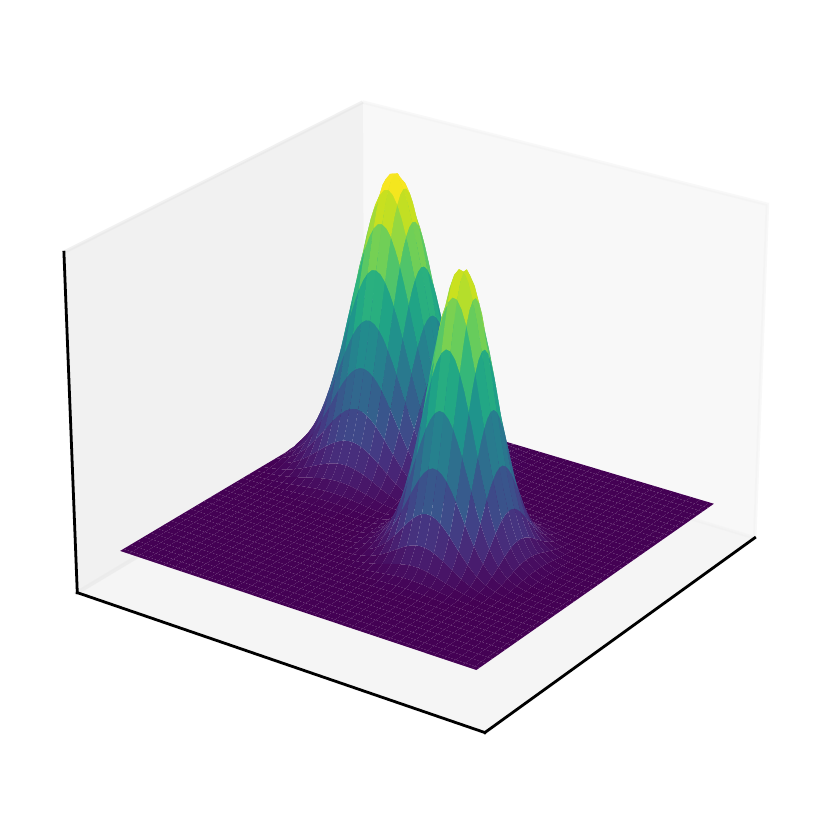}
                \end{subfigure}
            \end{subfigure}
            \begin{subfigure}{0.3\textwidth}
                \centering
                \begin{subfigure}[b]{0.55\textwidth}
                    \centering
                    \includegraphics[width=\textwidth]{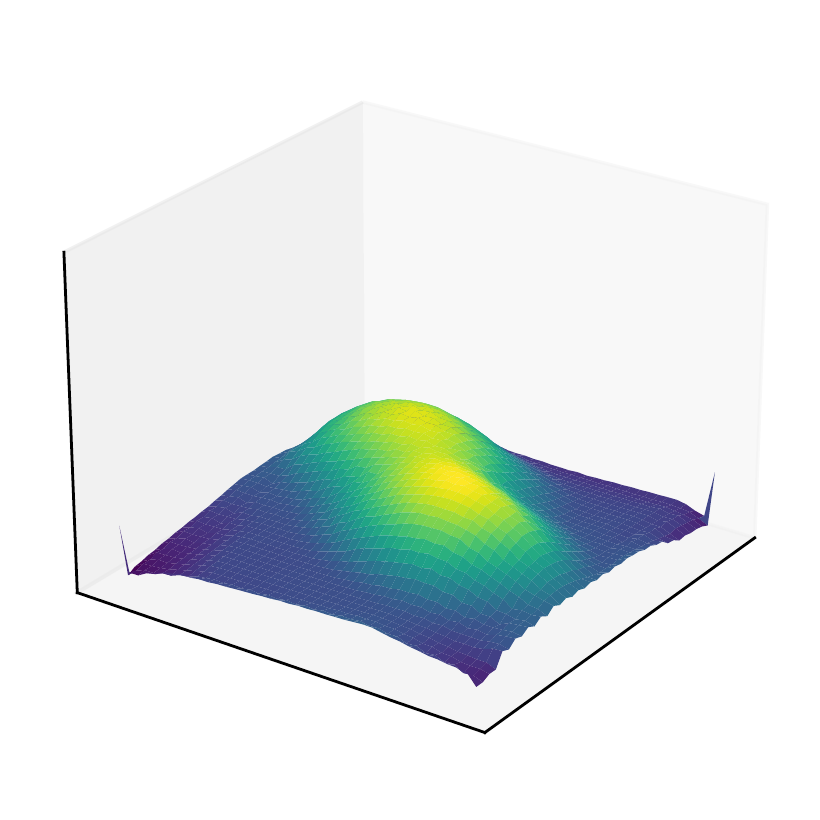}
                \end{subfigure}
                \begin{subfigure}[b]{0.4\textwidth}
                    \centering
                    \includegraphics[width=\textwidth]{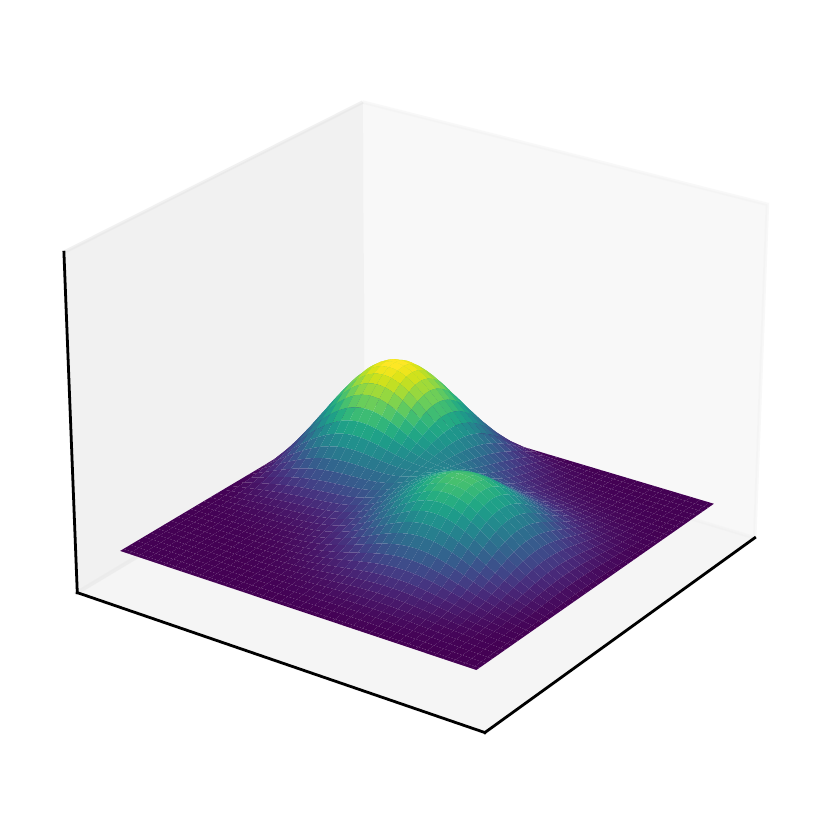}
                \end{subfigure}
            \end{subfigure}
            \begin{subfigure}{0.3\textwidth}
                \centering
                \begin{subfigure}[b]{0.55\textwidth}
                    \centering
                    \includegraphics[width=\textwidth]{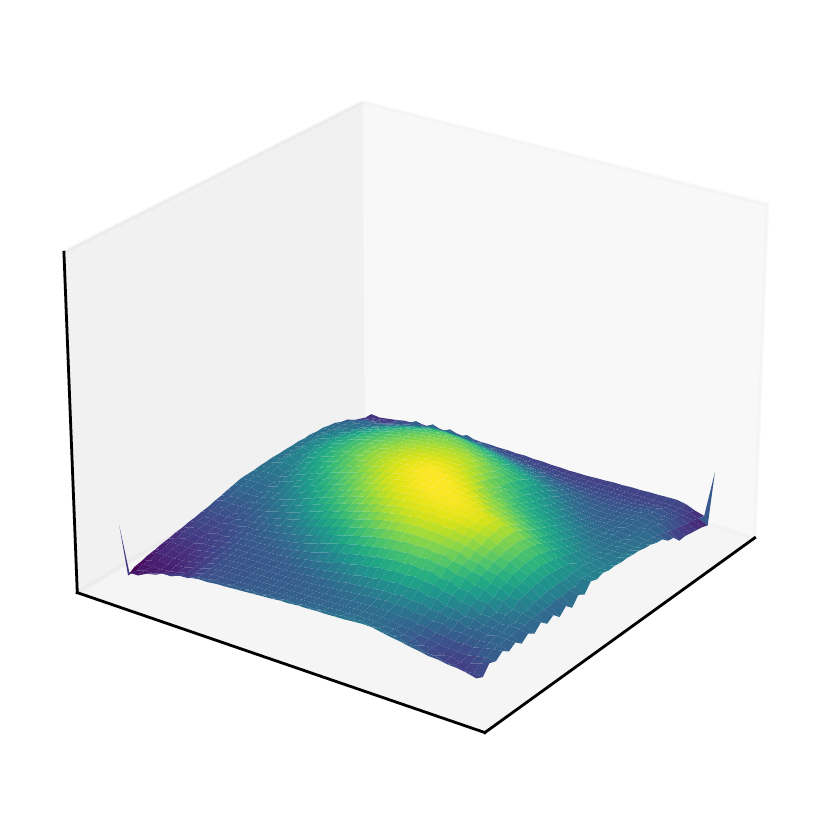}
                \end{subfigure}
                \begin{subfigure}[b]{0.4\textwidth}
                    \centering
                    \includegraphics[width=\textwidth]{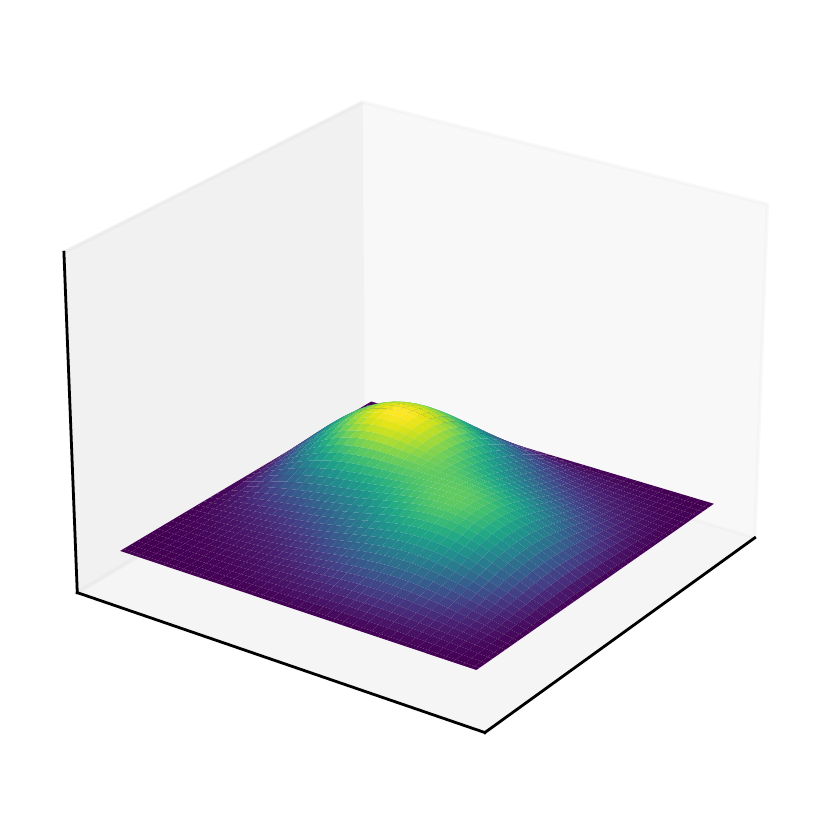}
                \end{subfigure}
            \end{subfigure}
        \end{subfigure}
    \end{subfigure}
    \\
    \begin{subfigure}{\textwidth}
        \centering
        \begin{subfigure}{\textwidth}
        \raisebox{0.0\textwidth}{\rotatebox[origin=t]{90}{\small{Initial Condition 3}}}
            \centering
            \begin{subfigure}{0.3\textwidth}
                \centering
                \begin{subfigure}[b]{0.55\textwidth}
                    \centering
                    \includegraphics[width=\textwidth]{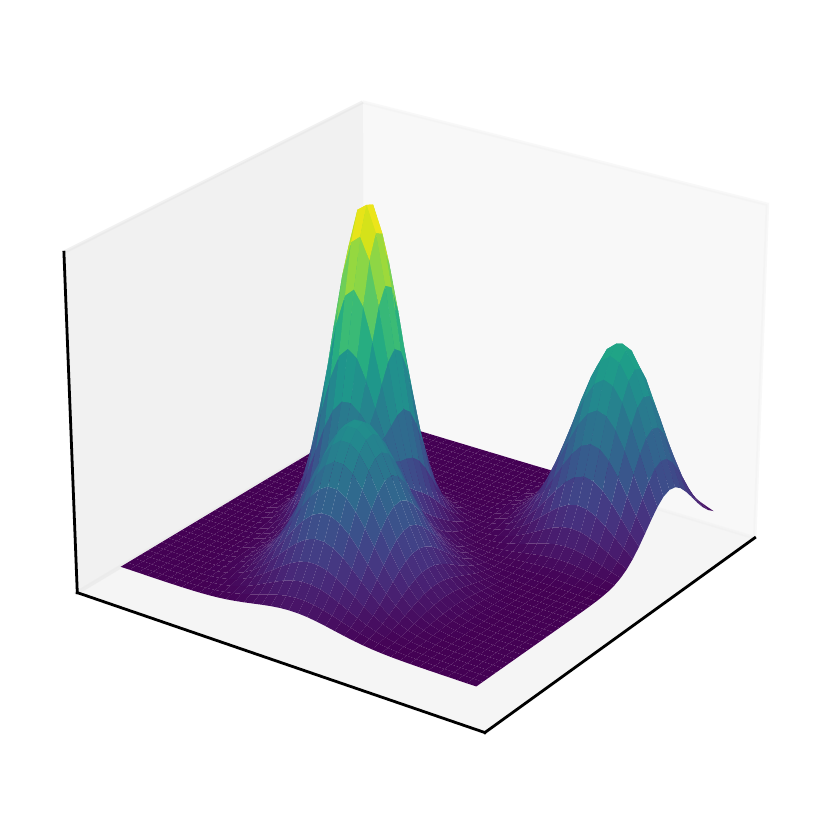}
                    \caption*{(Ours) \\ t = 0s}
                \end{subfigure}
                \begin{subfigure}[b]{0.4\textwidth}
                    \centering
                    \includegraphics[width=\textwidth]{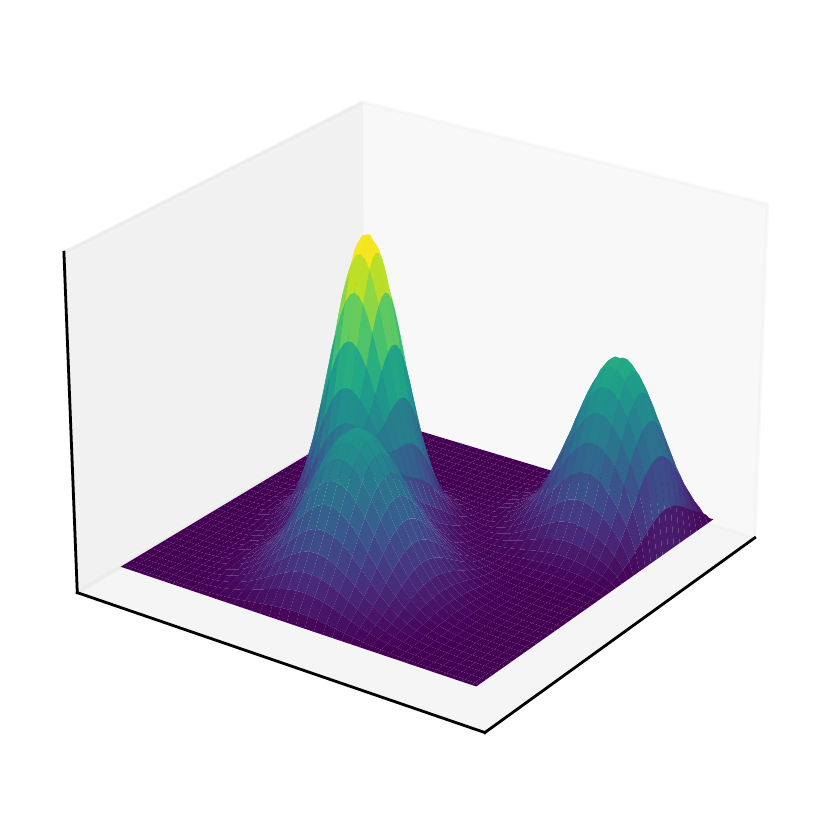}
                    \caption*{(Baseline)}
                \end{subfigure}
            \end{subfigure}
            \begin{subfigure}{0.3\textwidth}
                \centering
                \begin{subfigure}[b]{0.55\textwidth}
                    \centering
                    \includegraphics[width=\textwidth]{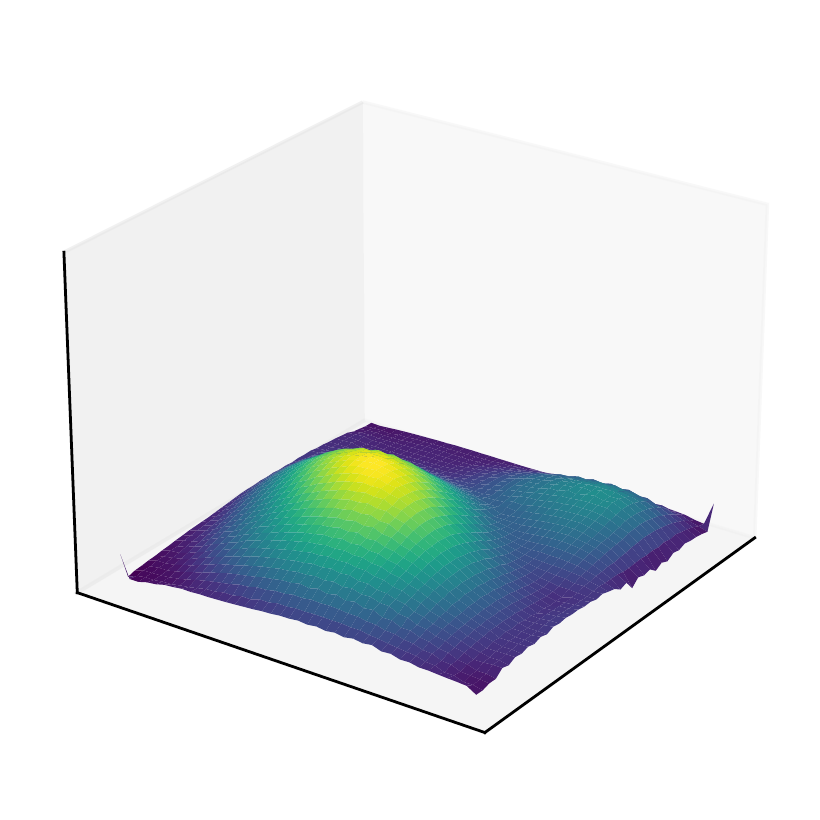}
                    \caption*{(Ours) \\ t = 0.1s}
                \end{subfigure}
                \begin{subfigure}[b]{0.4\textwidth}
                    \centering
                    \includegraphics[width=\textwidth]{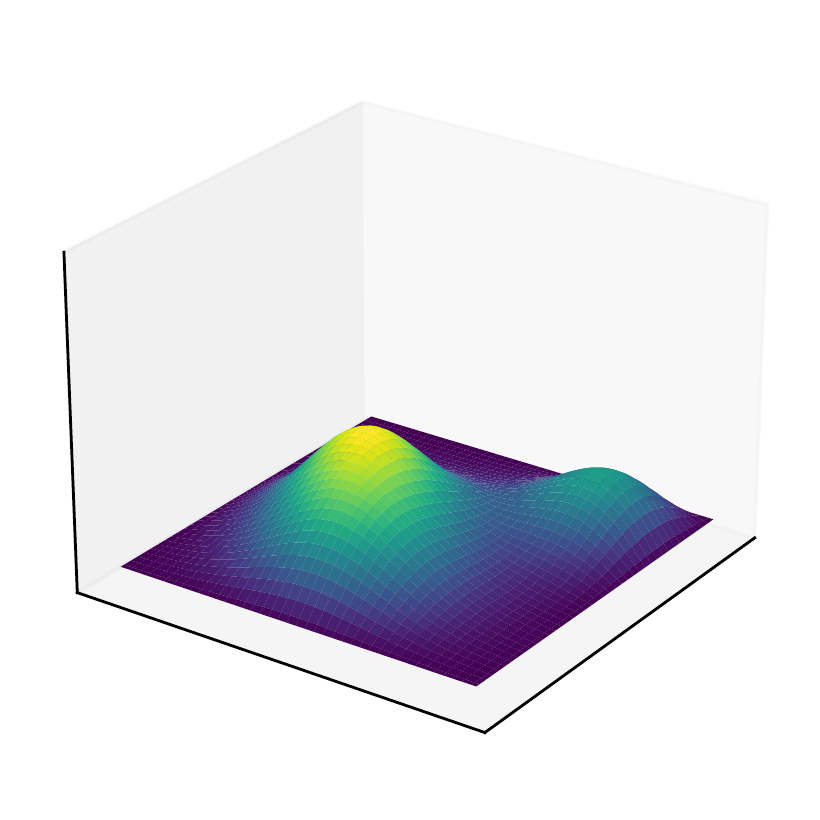}
                    \caption*{(Baseline)}
                \end{subfigure}
            \end{subfigure}
            \begin{subfigure}{0.3\textwidth}
                \centering
                \begin{subfigure}[b]{0.55\textwidth}
                    \centering
                    \includegraphics[width=\textwidth]{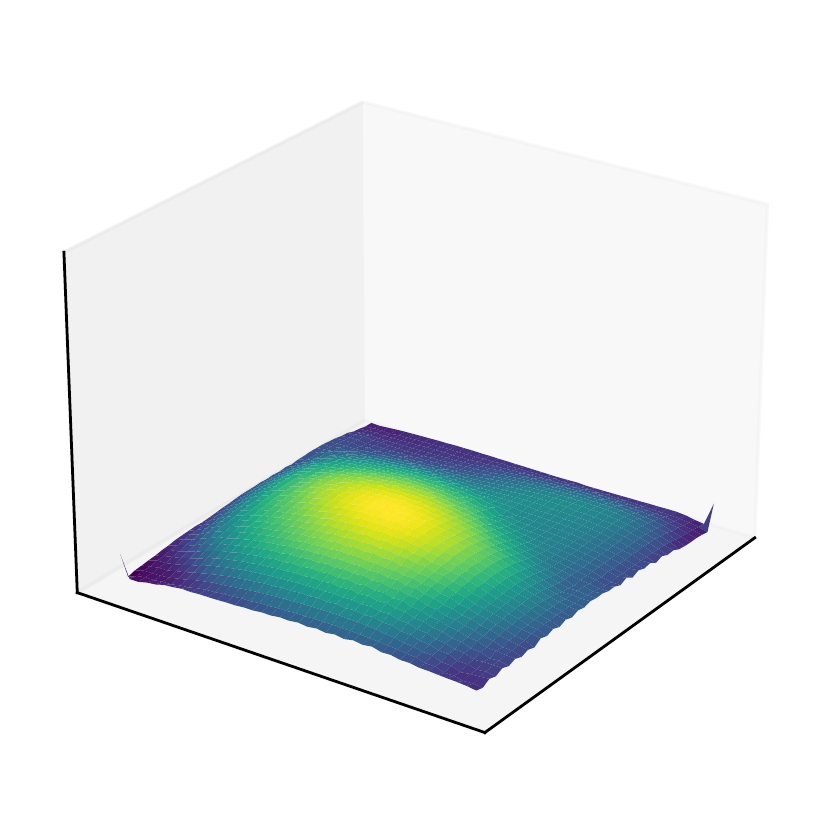}
                    \caption*{(Ours) \\ t = 0.2s}
                \end{subfigure}
                \begin{subfigure}[b]{0.4\textwidth}
                    \centering
                    \includegraphics[width=\textwidth]{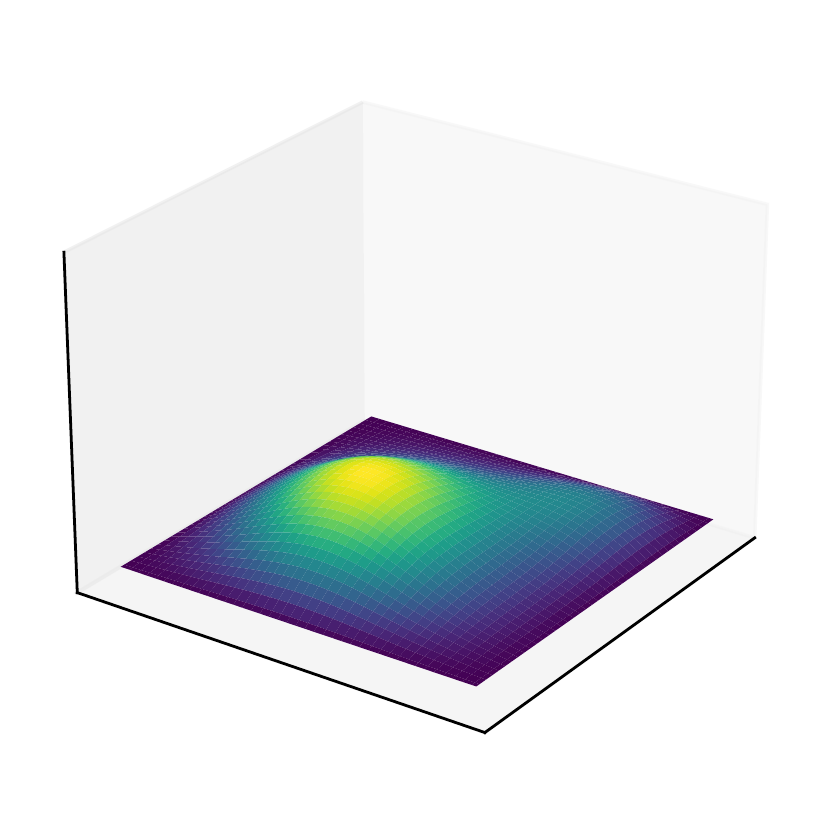}
                    \caption*{(Baseline)}
                \end{subfigure}
            \end{subfigure}
        \end{subfigure}
    \end{subfigure}
    
    \caption{Every row represents one random initial condition. Every horizontal pair of images is once the network prediction on the left and smaller baseline on the right. There are three pairs in every row where each pair represent the solutions at $t=0s$, $t=0.1s$, and $t=0.2s$, respectively.}
    \label{fig:heat2d}
\end{figure*}

\paragraph{Heat Equation 2D with Randomized Environment} In \cref{fig:heat2d} we show how the network performs on previously unseen initial conditions, predicting multiple timesteps in sequence. From \cref{table:benchmark} one can see from the Mean Squared Error (MSE) that it generalizes very well during testing. Our example scenario is still a rather simple one, but this shows the potential in having a generalizable, interactive solver for PDEs.

\paragraph{Benchmark against baseline} We compute the MSE between our network prediction and the FEniCS baseline and report inference times in \cref{table:benchmark}. Note that due to our problems being low-dimensional and having coarse meshes, the main overhead is the forward prediction iterations over all timesteps. The PDE problems are additionally also explicitly kept simple to verify fundamental properties of the physics-constrained learning method, hence we imagine in hard-to-solve practical scenarios such as Navier-Stokes equations, the speed-up will be far more significant. Nevertheless, we hope to provide a step in the right direction towards such physics-constrained unsupervised mesh learning methods.

These examples prove that we have a properly implemented the graph gradient, while also keeping a computationally light framework for gradient computation. We used FEniCS as groundtruth and showed that while our work produces qualitatively good results, the underlying limitation of finite difference will still need to be corrected in order to adopt this framework to complex physical phenomena.

\begin{table*}[!t]
\begin{minipage}{\textwidth}
    \centering
    \caption{Accuracy and inference times of our physics-constrained network compared with the FEniCS baseline in Python.}
    \label{table:benchmark}
    \vskip 0.15in
    \begin{center}
    \begin{small}
    \begin{sc}
    \begin{tabular}{lcccc}
    \toprule
    PDE & Mesh Nodes & Network Runtime & FEniCS Runtime & MSE\\
    \midrule
    Heat 1D     & 20\;000  & 325ms & 1.21s & 1.80e-3 \\
    Eikonal 1D  & 40\;000  & 323ms & 2.37s & 9.82e-4 \\ 
    Burgers 1D  & 100\;000 & 393ms & 1.28s & 1.05e-1 \\ 
    Heat 2D     & 160\;000 & 531ms & 3.06s & 6.94e-4 \\ 
    \bottomrule
    \end{tabular}
    \end{sc}
    \end{small}
    \end{center}
    \vskip -0.1in
    
\end{minipage}
\end{table*}

\section{Conclusion}
\label{sec:conc}
We present an unsupervised approach to training graph networks with knowledge of the governing physical equations. Previous methods have been relying on large amounts of pre-generated data to perform supervised learning until now. This work has moved to an unsupervised regime of physics-based machine learning, and unlike previous approaches on fixed uniform grids, we now learn on irregular meshes that can fit any geometry. We successfully utilized the previous success of graph networks for handling mesh dynamics, and have now brought those to an unsupervised setting.

A shortcoming of our current work is that our graph gradient breaks down due to the inaccuracies of finite differences for e.g. shock or rarefaction waves. Hence, we should look in the future towards learned gradients \citep{Seo2020Physics-aware,horie2021isometric} that allow an efficient and accurate computation, or implement a more accurate FEM-based approach that interpolates polynomials/basis functions as shown in \citep{wandel2021spline}.

Some may also consider discretizing a continuous space using a mesh as a limitation, since a lot of time and effort goes into the approximation of real-world objects as meshes, and works such as Physics-Informed Neural Networks (PINN) \citep{raissi2019physics} address this issue. There have additionally also been works focusing on making PINNs easier to train and more generalizable \citep{li2021physics}. However, how to apply these PINNs to real-world problems remains an open question, whereas object meshes are already used in a variety of real-world applications.





To better mimic the success achieved with U-Nets \citep{ronneberger2015u} on image-based inputs as shown in \citep{wandel2020learning}, we will in the future extend our graph networks to graph U-Nets \citep{gao2019graph}. That change would incorporate multi-scale resolution describing the governing physics at different length scales. 

As shown in \citet{pfaff2021learning} for supervised training, we don't have to be limited to only Eulerian meshes, but we can also apply unsupervised learning for Lagrangian meshes, which was previously not even possible on just fixed grids. Hence large hyperelastic deformations would be a future application for our physics-constrained, unsupervised learning framework for meshes.



\clearpage

\balance
\bibliography{biblio.bib}
\bibliographystyle{icml2022}

\newpage
\appendix
\onecolumn

\section{Derivation of Directional Graph Gradient}
\label{app:gradder}

In the case of linear triangular elements we model an interpolation matrix $\vec{H}$ for arbitrary values $u, v$ at each node. This matrix is also the transformation from natural to global coordinates (we will later want to compute the inverse):

\begin{equation}
\begin{aligned}
    \vec{H} &= \begin{pmatrix}
    1-r-s & 0 & r & 0 & s & 0 \\
    0 & 1-r-s & 0 & r & 0 & s
    \end{pmatrix}
    \\
    \begin{pmatrix} x(r,s) \\ y(r,s) \end{pmatrix} &= \vec{H} \begin{pmatrix} x_1 \\ y_1 \\ x_2 \\ y_2 \\ x_3 \\ y_3 \end{pmatrix} = \begin{pmatrix} x_1 + (x_2 - x_1) r + (x_3 - x_1) s \\ y_1 + (y_2 - y_1) r + (y_3 - y_1) s \end{pmatrix}
    \\
    \begin{pmatrix} u(r,s) \\ v(r,s) \end{pmatrix} &= \vec{H} \begin{pmatrix} u_1 \\ v_1 \\ u_2 \\ v_2 \\ u_3 \\ v_3 \end{pmatrix} = \begin{pmatrix} u_1 + (u_2 - u_1) r + (u_3 - u_1) s \\ v_1 + (v_2 - v_1) r + (v_3 - v_1) s \end{pmatrix}
\end{aligned}
\end{equation}

From this we can easily compute the Jacobian w.r.t. local coordinates (for some arbitrary value $u$):
\begin{align}
    \vec{J_{rs}} &= \begin{pmatrix}
    \frac{\partial u}{\partial r} & \frac{\partial u}{\partial s} \\
    \frac{\partial v}{\partial r} & \frac{\partial v}{\partial s}
    \end{pmatrix}
    =
    \begin{pmatrix}
    u_2 - u_1 & u_3 - u_1 \\
    v_2 - v_1 & v_3 - v_1
    \end{pmatrix}
\label{eq:trijac}
\end{align}

Given a value $u$ that varies throughout the finite element, we can write its total derivative with the chain rule $\frac{\partial u}{\partial r} = \frac{\partial u}{\partial x} \frac{\partial x}{\partial r} + \frac{\partial u}{\partial y} \frac{\partial y}{\partial r}$. In both dimensions this results in:
\begin{align*}
    \begin{pmatrix} \frac{\partial u}{\partial r} \\ \frac{\partial u}{\partial s} \end{pmatrix}  
    &= \begin{pmatrix} 
        \frac{\partial u}{\partial x} \frac{\partial x}{\partial r} + \frac{\partial u}{\partial y} \frac{\partial y}{\partial r} \\ 
        \frac{\partial u}{\partial x} \frac{\partial x}{\partial s} + \frac{\partial u}{\partial y} \frac{\partial y}{\partial s} 
    \end{pmatrix}  
    = \begin{pmatrix}
        \frac{\partial x}{\partial r} & \frac{\partial y}{\partial r} \\
        \frac{\partial x}{\partial s} & \frac{\partial y}{\partial s}
    \end{pmatrix}
    \begin{pmatrix} \frac{\partial u}{\partial x} \\ \frac{\partial u}{\partial y} \end{pmatrix} 
    = \begin{pmatrix}
        x_2 - x_1 & y_2 - y_1 \\
        x_3 - x_1 & y_3 - y_1
    \end{pmatrix}
    \begin{pmatrix} \frac{\partial u}{\partial x} \\ \frac{\partial u}{\partial y} \end{pmatrix} 
\end{align*}

Let us define $\vec{J} := \begin{pmatrix} \frac{\partial x}{\partial r} & \frac{\partial y}{\partial r} \\ \frac{\partial x}{\partial s} & \frac{\partial y}{\partial s} \end{pmatrix}$, which we can compute using the known positions of the nodes. Our initial objective is to calculate the derivatives $\frac{\partial u}{\partial x}$ and $\frac{\partial u}{\partial y}$ w.r.t. the global coordinates $x$ and $y$, which we can find with:
\begin{equation}
    \nabla u = \begin{pmatrix} \frac{\partial u}{\partial x} \\ \frac{\partial u}{\partial y} \end{pmatrix}  = \vec{J}^{-1} \begin{pmatrix} \frac{\partial u}{\partial r} \\ \frac{\partial u}{\partial s} \end{pmatrix} = 
    \vec{J}^{-1} \begin{pmatrix} u_2 - u_1 \\ u_3 - u_1 \end{pmatrix} 
\end{equation}
This matrix $\vec{J}$ is always invertible as long as the area of the triangular element is nonzero (then $\det{J} \neq 0)$. Note that this $\vec{J}$ is the transposed version of the Jacobian defined in~\eqref{eq:trijac} for position.

This gradient is constant throughout the triangular element. For the gradient values at each node of the mesh, we will average the gradients of all connecting elements. We achieve this averaging operation as follows: We first compute a binary adjacency matrix $\vec{A}$ to figure out which nodes are adjacent to which finite elements. Here $A_{ij} = 1$ if node $j$ is part of element $i$. Next we transpose this adjacency matrix, such that we have a new matrix $\vec{E} = \vec{A}^T$ where $E_{ij} = 1$ if element $j$ is adjacent to node $i$. If we multiply $\vec{E}$ with a vector containing values at every element, we will end up summing these values together and creating a vector of values at every node. Instead of summing, we want to take an average, hence we first compute the row-wise sums of $\vec{E}$, then divide every row by their sum. We implemented this by multiplying the diagonalized row-sum vector with $\vec{E}$. This resulting matrix (which can be pre-computed) can be multiplied with the previously computed gradients at every element to retrieve the gradient values at every node in the graph.

\section{1D Spatial Gradient}
\label{app:grad1D}

For simple PDEs (such as Eikonal or Burger's equation) we often have a single spatial dimension, hence we will describe a short extension to the previous method for this 1D scenario. We once again describe global line segments in local coordinates, where the segment/graph edge has 2 nodes (globally at $x_1$ and $x_2$) that start at 0 and end at 1 (1D coordinates).
\begin{align}
    x(r) &= \begin{bmatrix} 1 - r &  r \end{bmatrix} \begin{bmatrix} x_1 \\ x_2 \end{bmatrix} = x_1 + (x_2 - x_1) r 
\end{align}
Computing derivatives in this case is as simple as:
\begin{align}
    \frac{\partial u}{\partial x} = \frac{\partial r}{\partial x} \frac{\partial u}{\partial r} = \left( \frac{\partial x}{\partial r} \right)^{-1} \frac{\partial u}{\partial r} 
    = \frac{1}{\left( x_2 - x_1 \right)} \frac{\partial u}{\partial r}
\end{align}

\section{3D Spatial Gradient}
\label{app:grad3D}

Similarly to the previous approaches, we might want the gradient in 3D in case we're dealing, for example, with Lagrangian meshes for solid deformations. We define the tetrahedral element in natural/local coordinates with the following 4 vertices $\vec{x}_0, \vec{x}_1, \vec{x}_2, \vec{x}_3$ respectively: $(0,0,0), (1,0,0), (0,1,0), (0,0,1)$. We find the following Jacobian:
\begin{align}
    \vec{J} := \begin{pmatrix} 
        \frac{\partial x}{\partial r} & \frac{\partial y}{\partial r} & \frac{\partial z}{\partial r} \\ 
        \frac{\partial x}{\partial s} & \frac{\partial y}{\partial s} & \frac{\partial z}{\partial s} \\
        \frac{\partial x}{\partial t} & \frac{\partial y}{\partial t} & \frac{\partial z}{\partial t}
    \end{pmatrix}
    = \begin{pmatrix} 
        x_1 - x_0 & y_1 - y_0 & z_1 - z_0 \\ 
        x_2 - x_0 & y_2 - y_0 & z_2 - z_0 \\
        x_3 - x_0 & y_3 - y_0 & z_3 - z_0
    \end{pmatrix}
\end{align}

Finding the gradient of some value $u$ on the mesh now simplifies to:
\begin{equation}
    \nabla u = \begin{pmatrix} \frac{\partial u}{\partial x} \\ \frac{\partial u}{\partial y} \\ \frac{\partial u}{\partial z} \end{pmatrix}  = \vec{J}^{-1} \begin{pmatrix} \frac{\partial u}{\partial r} \\ \frac{\partial u}{\partial s} \\ \frac{\partial u}{\partial t} \end{pmatrix} = 
    \vec{J}^{-1} \begin{pmatrix} u_1 - u_0 \\ u_2 - u_0 \\ u_3 - u_0 \end{pmatrix} 
\end{equation}



\end{document}